\documentclass[twoside,11pt]{article}

\usepackage{jagi}
\usepackage{times}

\usepackage[titlenumbered,ruled,linesnumbered]{algorithm2e}
\DontPrintSemicolon
\usepackage{wrapfig}

\usepackage{booktabs}       
\usepackage{amsmath}
\usepackage{subcaption}
\usepackage{wrapfig}
\usepackage{color,soul}
\usepackage{enumitem}
\usepackage[dvipsnames]{xcolor}
\usepackage{wasysym}

\usepackage{floatrow}
\newfloatcommand{capbtabbox}{table}[][\FBwidth]

\usepackage{hyperref}

\usepackage{xspace}
\newcommand{\LLL}{LML\xspace}

\firstpageno{1}

\begin{document}

\title{A Deep Learning Framework for Lifelong Machine Learning}

\author{\name Charles X. Ling \email charles.ling@uwo.ca \\
       \name Tanner Bohn \email tbohn@uwo.ca \\
       \addr Western University \\
       1151 Richmond St. \\
       London ON, N6A 3K7, Canada}


\maketitle

\begin{abstract}
Humans can learn a variety of concepts and skills incrementally over the course of their lives while exhibiting many desirable properties, such as continual learning without forgetting, forward transfer and backward transfer of knowledge, and learning a new concept or task with only a few examples.  Several lines of machine learning research, such as lifelong machine learning, few-shot learning, and transfer learning attempt to capture these properties. However, most previous approaches can only demonstrate subsets of these properties, often by different complex mechanisms. In this work, we propose a simple yet powerful unified deep learning framework that supports almost all of these properties and approaches through {\em one} central mechanism.  
Experiments on toy examples support our claims. We also draw connections between many peculiarities of human learning (such as memory loss and ``rain man'') and our framework. 

As academics, we often lack resources required to build and train, deep neural networks with billions of parameters on hundreds of TPUs. 
Thus, while our framework is still conceptual, and our experiment results are surely not SOTA, we hope that this unified lifelong learning framework inspires new work towards large-scale experiments and understanding human learning in general.  

This paper is summarized in two short YouTube videos: \url{https://youtu.be/gCuUyGETbTU} (part 1) and \url{https://youtu.be/XsaGI01b-1o} (part 2).

\end{abstract}

\begin{keywords}
  Lifelong machine learning, Multi-task learning, Deep learning, Human-inspired learning, Weight consolidation, Neural networks
\end{keywords}

\section{Introduction}
\label{sec:introduction}

The past decade has seen significant growth in the capabilities of artificial intelligence and machine learning (ML).
Deep learning in particular has archived great successes in medical image recognition and diagnostics \citep{litjens2017survey, shen2017deep}, tasks on natural language processing \citep{radford2019language, devlinetal2019bert}, difficult games \citep{silver2017mastering}, and even farming \citep{kamilaris2018deep}.
However, deep learning models almost always need thousands or millions of training samples, given at once to train the models, to perform well.
This is in a sharp contrast with human learning, which normally learns a new concept with a small number of samples, and continues to learn over a lifetime. 
Other major weaknesses in current deep learning, when compared to human learning, include 
difficulty in leveraging previous learned knowledge to better learn new concepts (and vice versa) and 
learning many tasks sequentially without forgetting previous ones.

Several lines of research in supervised learning exist to overcome these weaknesses.
Multi-task learning \citep{caruana1997multitask} considers how to learn multiple concepts at the same time such that they help each other to be learned better.
The related field of transfer learning \citep{pan2009survey} assumes that some concepts have been previously learned and we would like to transfer their knowledge to assist learning new concepts.
Few-shot learning \citep{fei2006one} aims to learn tasks with a small number of labeled data.  
Lifelong machine learning (\LLL) \citep{thrun1998lifelong, LLL}, also known as continual \citep{parisi2019continual} or sequential learning \citep{CatastrophicForgetting}, considers how to learn and transfer skills across long sequences of tasks.

However, most previous \LLL approaches only demonstrate subsets of these human-like properties by different complex mechanisms. It is our belief that human lifelong concept learning with labeled examples likely uses a single (or small set of) mechanism(s).  This is because humans learn many concepts in their lives, and the process is {\em continuous without sharp boundary}. 
For example, learning a new concept (such as a new shape) or updating a learned
concept (such as receiving more data of a previously learned shape) may subtly influence concepts learned before and after.  Using \LLL terminology, this subtle influence is due to forward and backward transfer, non-forgetting of some previous concepts, and graceful forgetting of others. 
These influences appear to combine seamlessly in humans, providing effective lifelong learning capabilities. 
Thus, we wish to find a learning framework with one central mechanism that can seamlessly implement such influences\footnote{More specifically, we are aiming to describe a higher-level \textit{algorithmic} mechanism, rather than provide a model of specific neural processes associated with human learning.}, and demonstrate many human-like lifelong learning properties.

\begin{figure}[ht]
    \centering
    \includegraphics[width=0.75\textwidth]{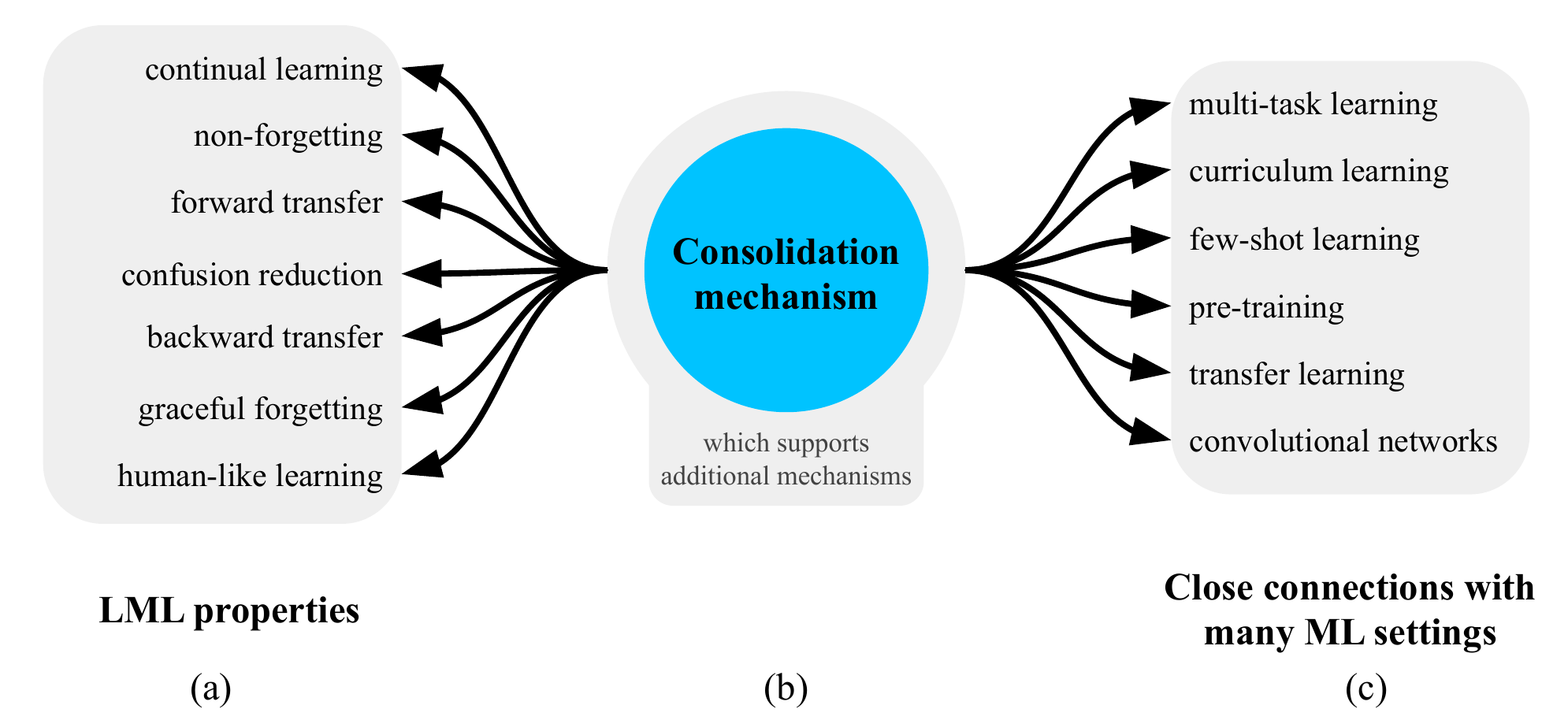}
    \caption{An overview of our unified framework: by combining a central consolidation mechanism with additional tools, we are able to exhibit many lifelong learning properties and demonstrate close connections with other ML settings. The set of desirable \LLL properties in (a) is discussed in Section~\ref{sec:properties}. A description of the consolidation mechanism and its combination with the additional mechanisms shown in (b) is discussed in Section~\ref{sec:framework}. 
    The learning settings listed in (c) where our framework can be applied is discussed in Section~\ref{sec:discussion}. 
    The connection between our framework and human learning in particular is expanded upon in Section~\ref{sec:parallels}.
    }
    \label{fig:framework}
\end{figure}  

In this paper, we propose such a 
\textit{unified framework with one central mechanism} with the support of other mechanisms such as network expansion and rehearsal. A high-level overview of the framework is depicted in Figure~\ref{fig:framework}.
We demonstrate its unified characteristic by discussing how it can illustrate many desirable \LLL properties such as non-forgetting, forward and backward transfer, and graceful forgetting (Section~\ref{sec:framework}). To support our framework, we include proof-of-concept experimental results for our \LLL properties in Section~\ref{sec:experiments}.
We also discuss connections to other ML settings, 
and draw parallels with many peculiarities seen in human learning, such as memory loss and ``rain man'' in Section~\ref{sec:parallels}. 
We hope that this perspective can shed new light on human learning. 

Our paper is different from most deep learning papers that aim to achieve state-of-the-art results. 
Instead, our paper shows the generality of a new framework that can exhibit many properties, apply to many ML settings, and encompass previous works.
We hope it inspires more researchers to engage in the various related topics towards better understanding of ML, human general intelligence, and human learning.

\section{Lifelong Learning and its Properties}
\label{sec:overview}

In this section we will first describe our \LLL setting and its relation to other learning settings. We will then discuss a broad set of important \LLL properties. Lastly, we will provide a comparison of common approaches to \LLL.

\subsection{Lifelong Learning Setting}
\label{sec:setting}



In our lifelong setting, we mainly consider the \textit{task-incremental} classification tasks, where batches of data for new tasks arrive sequentially.
That is, a sequence of $(T_{0}, D_{0}), (T_{1}, D_{1}), ...$ are given, where $D_{i}$ is the labeled training data of task $T_{i}$.  
Classification models for $(T_{0}, T_{1}, ..., T_{j})$
must be learned and functional before $(T_{j+1}, D_{j+1})$ arrives. 
This models the incremental process of human lifelong learning. 

\begin{figure}[ht]
    \centering
    \includegraphics[width=1\textwidth]{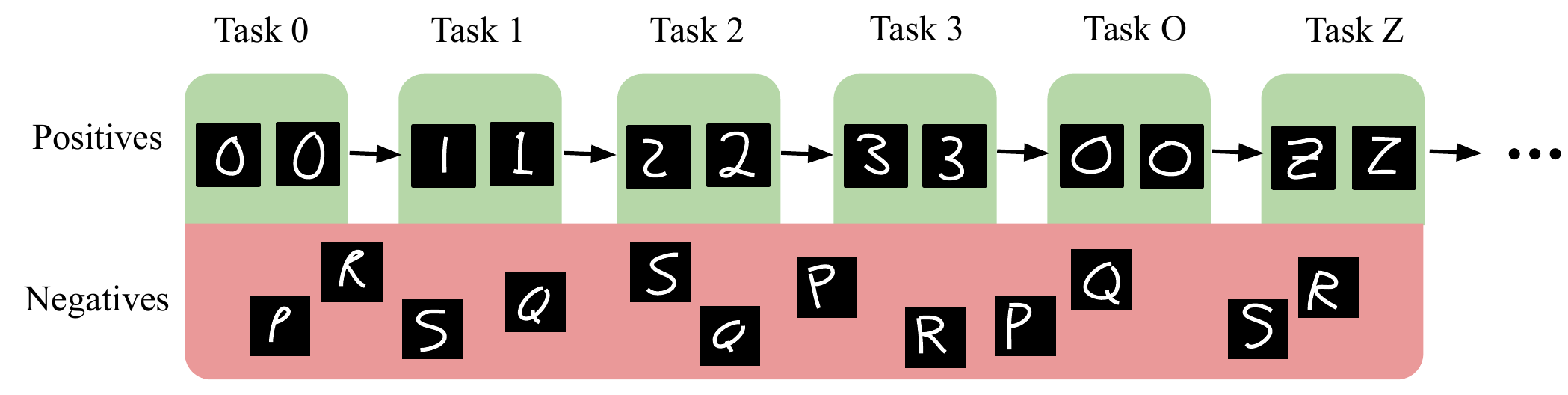}
    \caption{The sequence of binary classification tasks used in our 
    paper as a running example.  Here we assume that the same group of negative characters is shared across tasks\protect\footnotemark.}
    \label{fig:task_seq}
\end{figure}

As a toy running example for this paper, we assume that the sequence of the classification tasks is to learn to classify hand-written numbers and letters, starting with $T_{0} =$ ``0'', then $T_{1} =$ ``1'', then $T_{2} =$ ``2'', and so on, as seen in Figure~\ref{fig:task_seq}. It will be clear soon why we use hand-written letters ``O'' and ``Z'' in our running example.  

\footnotetext{Also worth considering for this toy example is where negative examples of each task include some positive samples of the tasks that came \textit{before} (since this data will have been made available).}

Our \LLL setting is distinct from most supervised learning and multi-task learning settings in two important ways:
\begin{enumerate}
    \item The data for task $T_{j}$ is only available when learning $T_{j}$ (and not before). Similarly, data for tasks $T_{j+1}$ and so on are not yet available, but classifiers for Tasks $T_{0}$ to $T_{j}$ must be built.  That is, one cannot wait until all data is available to build classifiers, as in the ``batch mode'' of learning.
    \item When building a new classifier for $T_{j}$, the \LLL algorithm should refrain, as much as possible, from using data from previous tasks; otherwise, when learning the last task using data of all previous tasks, it becomes the traditional ``batch mode'' again. 
\end{enumerate}
We believe that our lifelong learning setting closely matches how humans learn concepts in sequence in their lifetime. 

\begin{figure}[ht]
    \centering
    \includegraphics[width=0.75\textwidth]{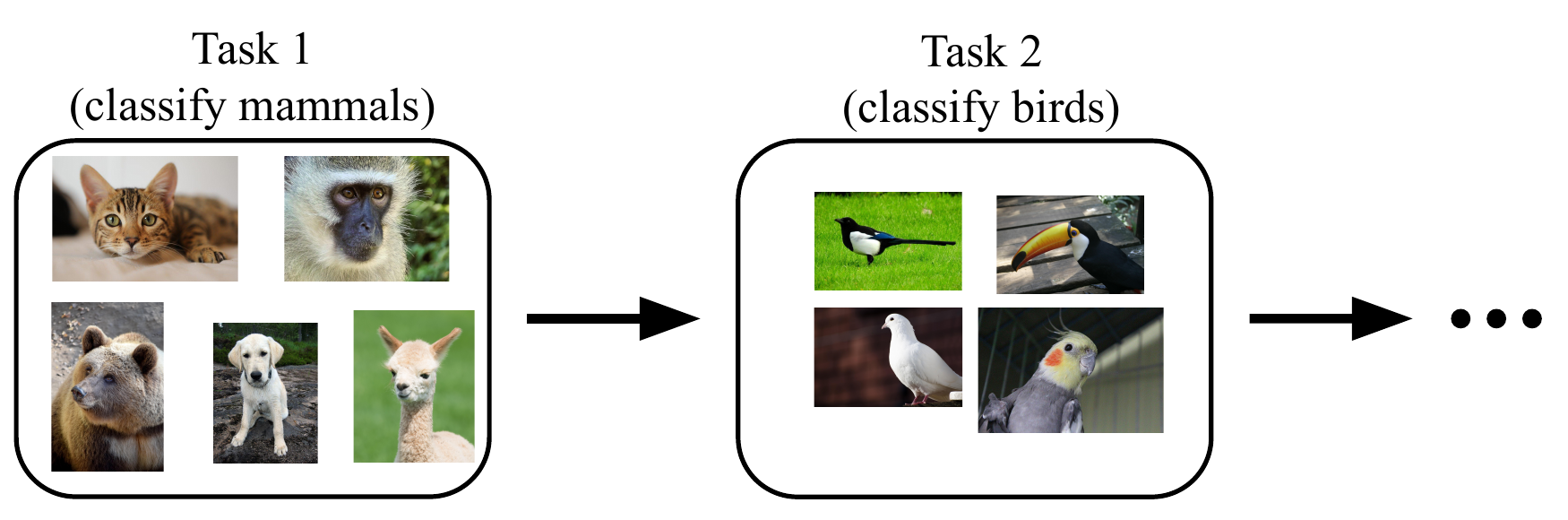}
    \caption{An example sequence of multi-class tasks, first learning to classify various mammals and then birds.}
    \label{fig:multiclass_seq}
\end{figure}

Our \LLL setting to learn one concept at a time can be easily extended to learning multi-class classifiers in sequence.  For example, the first task can learn to classify different mammals (given data of mammals), and then learn to classify various birds (given data of birds), and so on.  This is illustrated in Figure~\ref{fig:multiclass_seq}.
Another direction of extension is ``anytime'' \LLL, in which data of any task might be received at any time, and the relevant classifier must be updated with minimal use of data and interference on other learned classifiers. 

\subsection{Lifelong Learning Properties}
\label{sec:properties}

Here we discuss at a high level several properties a \LLL approach would ideally exhibit.
We will use the running example from Figure~\ref{fig:task_seq} of learning a sequence of binary classifiers for ``0'', ``1'', ``2'', ``3'', ``O'', and ``Z''.

\paragraph{Continual learning and deploying} 
As mentioned earlier, before starting to learn a new task, $T_{j}$, our \LLL approach should be able to perform well on all earlier tasks, $T_{<j}$. 
The data, $D_j$, for $T_{j}$, is only available when learning $T_{j}$.
While learning $T_{j}$, \LLL should refrain from using the previous task data, $D_{<j}$. 
This is in contrast to standard multi-task (batch) learning, where all data of all tasks are used for training at the same time. This continual learning condition ensures that the model is 1) useful, since each task must be learned to an acceptable performance level whenever data is available, 2) flexible, in that new tasks can be continually accommodated, 3) efficient, in that tasks are learned with high computational and data efficiency, and 4) human-like, in that humans seem to learn in a similar continuous fashion over their lifetime.  See Section~\ref{sec:continual} for details.


\paragraph{Non-forgetting} This is the ability to avoid catastrophic forgetting \citep{CatastrophicForgetting}, where learning $T_{j}$ causes a dramatic loss in performance on $T_{<j}$. Ideally, learning $T_{j}$ when using only the data of $T_{j}$ would not negatively affect $T_{<j}$. 
For example, learning the classifier for $T_{1}$ of ``1'' should not cause performance on the previous task of ``0'' to degrade, when the classifier for task ``0'' is not trained at that time. See Section~\ref{sec:continual} for details.


\paragraph{Forward transfer} This is the ability to learn new tasks, $T_{\geq j}$, easier and better following earlier learned similar tasks, $T_{<j}$.  This is also known as knowledge transfer \citep{pan2009survey}. 
Achieving sufficient positive forward transfer also enables \textbf{few-shot learning} of later concepts.
For example, first learning to classify ``0'' should allow the later task of ``O'' to be learned faster, as they are very visually similar.  
On the other hand, transferring between non-similar tasks may lead to negative transfer, compromising the performance of the new task. See Section~\ref{sec:forwardtransfer} for details.


\paragraph{Confusion reduction}
%
Classification algorithms often find the minimal set of discriminating features necessary for learning the tasks at hand.  
As training of each new task is performed individually with its training data, the testing accuracy can be high when testing on its own test data. 
However, when tested on data of all tasks, confusions can happen.  
For example, when learning the sequence of tasks of ``0'', ``1'', ``2'', ``3'', ``O'', and ``Z'', the two tasks of ``0'' and ``O'' may be confused (images of ``0'' may be predicted as ``O'', and vice versa). Similarly for ``2'' and ``Z''.  In such cases, we will need to resolve confusion between pairs of similar tasks. 


In human lifelong learning, this type of confusion may happen too. For example, when meeting new people, if the first two people are visually distinct (such as very tall vs. very short) we can rely only on this feature to tell them apart. However, if more people arrive and they are similar to the first two people, we may initially confuse them and must find finer details to reduce confusion, or to uniquely distinguish them. In the extreme case where we encounter identical twins, significant effort may be required to learn the necessary details (by re-using their facial image data) to resolve confusion. See Section~\ref{sec:confusionreduction} for details.

\paragraph{Graceful forgetting} 
A valuable property opposite to non-forgetting is \textit{graceful} forgetting \citep{aljundi2018memory}, often seen in humans. In our framework, learning new tasks requires additional model capacity, and when sufficient expansion is not possible, the model can turn to graceful forgetting of unimportant tasks to free up capacity for new tasks. See Section~\ref{sec:continual} and \ref{sec:gracefulforgetting} for details.

\paragraph{Backward transfer} This is knowledge transfer from $T_{\geq j}$ to $T_{<j}$, the opposite direction as forward transfer.   When learning a task, $T_{j}$, it may in turn help to improve the performance of $T_{< j}$.  This is like a ``review'' before a final exam after materials of all chapters have been taught and learned. Later materials can often help better understand earlier materials. See Section~\ref{sec:backwardtransfer} for details.

\paragraph{Human-like learning} As a new type of evaluation criteria for \LLL, we can consider how well an approach is predictive of human learning behaviour.
If a \LLL approach or framework is also able to provide explanatory power and match peculiarities of human learning (such as confusion or knowledge transfer in certain scenarios), it would have value in fields outside of ML. 
We discuss such connections between our framework and human learning in Section~\ref{sec:parallels}.

\subsection{Comparison of Different Lifelong Learning Approaches}
\label{sec:comparison}

The mechanisms used in previous work to perform \LLL tend to fall into three categories, and they can often 
only demonstrate subsets of \LLL properties as previously discussed. 
The first mechanism, replay, commonly works by storing previous task data and training on it alongside new task data \citep{rebuffi2017icarl, isele2018selective, chaudhry2019continual, wu2019large}. As a result of its data and computation inefficiency, we consider it generally not to be very a human-like learning mechanism. 
An exception to this is generative replay, where the storing of previous task samples is replaced by a deep generative model trained to produce samples for interleaving with new task samples. This approach, which requires a constant memory overhead is inspired by the generative ability of the primate hippocampus \citep{shin2017continual, kamra2017deep}.

The second mechanism is regularization. This mechanism works by restricting weight changes (making them less ``flexible'') via a loss function so that learning new tasks does not significantly affect previous task performance \citep{kirkpatrick2016EWC, zenke2017synaptic,chaudhry2018riemann,ritter2018online,li2017learning, zhang2020class}. We use this mechanism as an essential component in our unified framework. Compared to previous approaches, we propose to use regularization more strategically. Instead of simply controlling weight flexibility for non-forgetting, 
we leverage it to also encourage forward and backward transfer (Section~\ref{sec:forwardtransfer} and \ref{sec:backwardtransfer}), reduce confusion (Section~\ref{sec:confusionreduction}), and perform graceful forgetting (Section~\ref{sec:gracefulforgetting}).

The third mechanism, dynamic architecture, commonly works by adding new weights for each task and only allowing those to be tuned \citep{rusu2016progressive, yoon2018lifelong, xu2018reinforced}. This is often done without requiring previous task data and completely reduces forgetting while also allowing previous task knowledge to speed up learning of the new task. While this mechanism is necessary for \LLL of an arbitrarily long sequence of tasks (any fixed-size network will eventually reach maximum capacity), it should be used sparingly to avoid unnecessary computational costs. In Sections~\ref{sec:continual}, \ref{sec:forwardtransfer}, and \ref{sec:confusionreduction}, we describe how dynamic architectures can be efficiently used to help achieve many \LLL properties by combining it with our central mechanism.



\section{A Unified Framework for Lifelong Learning}
\label{sec:framework}

In this section we describe how our unified framework works. We start by introducing the central mechanism and in the rest of the section, discuss how to use this mechanism and combine it with additional mechanisms to achieve the several desirable \LLL properties described in Section~\ref{sec:properties}.

While not restricted to a particular neural network type, we primarily consider our framework as applied to deep neural networks, which have become popular in recent years, and are an attractive type of ML model due to their ability to automatically learn abstract features from data.

\subsection{A Central Consolidation Mechanism}

We propose a \LLL framework which situates a consolidation policy as the central mechanism. The consolidation policy works through a  hyperparameter, $\pmb{b}$, 
which controls the \textit{flexibility of
the model parameters} to be learned. When using neural networks (as considered in this paper), $\pmb{b}$ controls how flexibly the weights can be modified during gradient descent (or by any weight updating algorithm). In regression, $\pmb{b}$ controls how much coefficients can be modified. In rule-based learning, $\pmb{b}$ may control or regulate how much rule conditions can be added or deleted, or how much rule confidences can be updated, and so on. 
In this paper, we will mainly focus on deep neural networks. 


%
As deep learning essentially works through the minimization of a properly defined loss function, our 
consolidation mechanism essentially 
works as a regularization term in the loss function.
More specifically, 
if each network weight, $\theta_{i}$, is associated with a consolidation value of $\pmb{b}_{i} \geq 0$, the new loss function, $L_{new}$
while learning a given task of index $t$ is defined as follows:

\begin{equation}
    L_{new}(\theta) = L_{t}(\theta) + \sum_{i}\pmb{b}_{i}(\theta_{i} - \theta_{i}^{target})^{2}
\label{eqn:loss}
\end{equation}

Here, $\theta_{i}^{target}$ is the target value for a weight to be changed to. $L_{t}$ is a standard loss (such as cross-entropy) on task $t$.
This loss has the following behaviour: a large $\pmb{b}_{i}$ causes changing $\theta_{i}$ away from $\theta_{i}^{target}$ to be strongly penalized during training. When $\pmb{b}_{i} = \infty$, we refer to these weights as \textbf{``frozen''}, and simply fix them during training. 
In this case we can consider $\theta_{i}$ to be masked during backpropagation and completely prevented from changing to improve efficiency.
In contrast, $\pmb{b}_{i} = 0$ indicates that the weight is free to change, i.e. it is \textbf{``unfrozen''}.



\subsection{Continual Learning of New Tasks Without Forgetting}
\label{sec:continual}

In both lifelong and human learning, we desire to learn new tasks after learning previous tasks. In humans, this is supported by the ability to continually grow new connections between neurons and remove old connections \citep{cunha2010simple}. When this ability is compromised, so is our ability to learn new things. Similarly, in our conceptual framework we consider learning new tasks through the strategic and flexible use of network expansion. 

The pseudo-code in Algorithm~\ref{alg:continual} describes how to learn a new task, $T_{j}$, in a deep neural network in our conceptual framework after previous tasks, $T_{0},..., T_{j-1}$, have been learned.

\begin{algorithm}[h]
\SetAlgoLined
    \tcp{Given that tasks $T_{0}, ..., T_{j-1}$ have been learned}
    Recruit free units for $T_{j}$ and unfreeze new weights \tcp{Blue links in Figure~\ref{fig:continual}}
    Freeze weights of previous tasks \tcp{Red links in Figure~\ref{fig:continual} for non-forgetting}
    
    Initialize weights from earlier units to newly recruited units as described in Section~\ref{sec:forwardtransfer} \tcp{Green links in Figure~\ref{fig:continual} for forward transfer}
    
    Train the new task $T_{j}$ to minimize Eq.~\ref{eqn:loss}  \tcp{only on the data of new task $T_{j}$}
\caption{Continual Learning without Forgetting}
\label{alg:continual}
\end{algorithm}

\begin{figure}[ht]
    \centering
    \includegraphics[width=0.9\textwidth]{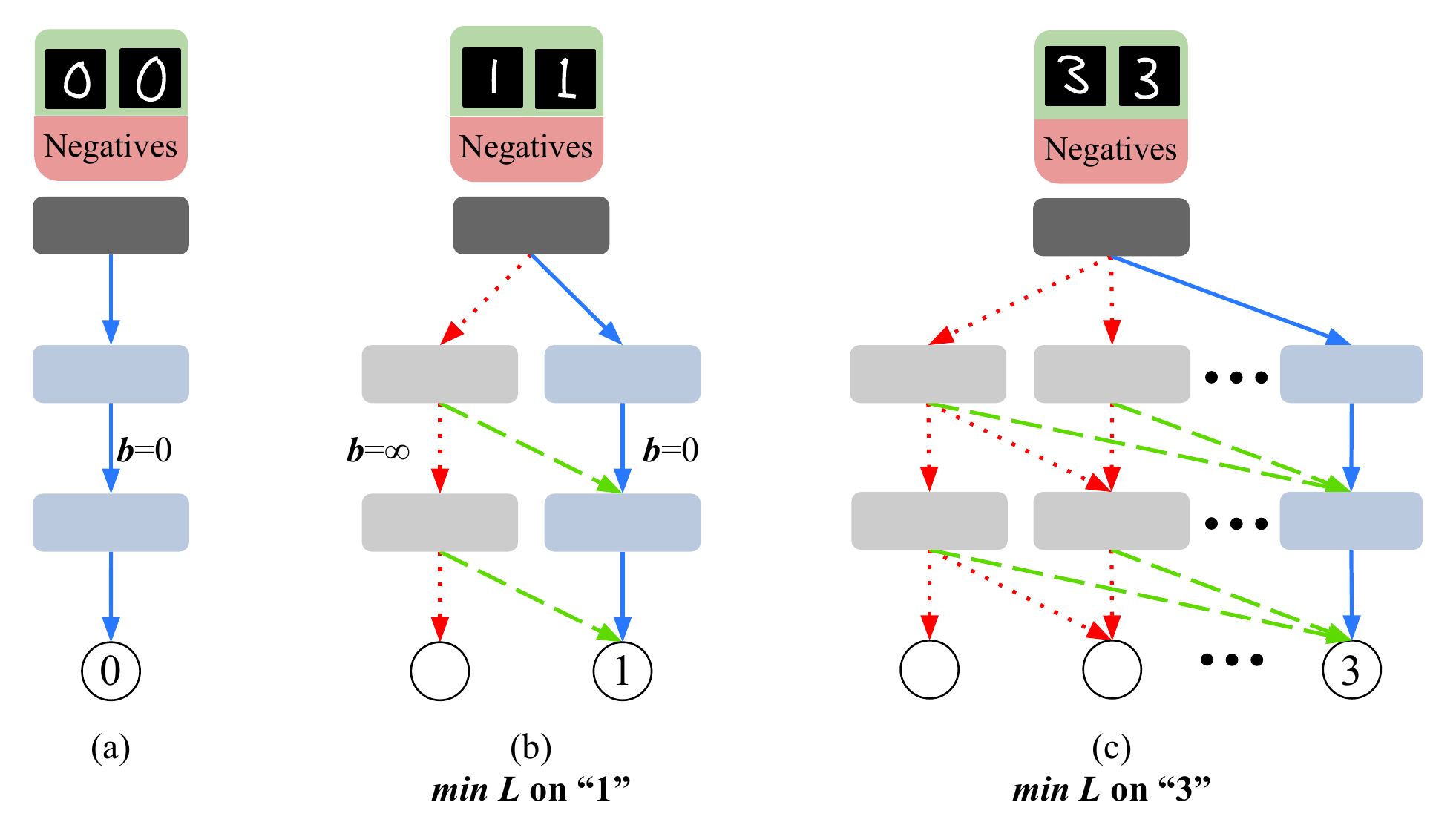}
    \caption{Continual learning with network expansion and without forgetting.   In (a), $T_{0}$ is being trained. All blue weights are randomly initialized and free to tune ($\pmb{b} = 0$).  In (b), $T_{1}$ is being learned without forgetting $T_{0}$, by freezing weights for $T_{0}$ (red links; $\pmb{b} = \infty$).  Note that the loss function is minimized only for the ``1'' classifier and data for ``0'' is not needed.  Here, green links are for forward transfer (see Section~\ref{sec:forwardtransfer}), and blue links are free to tune.  In (c), $T_{3}$ is learned without forgetting of the previous tasks.  The loss function is minimized only for the ``3'' classifier.}
    \label{fig:continual}
\end{figure}

As seen in Figure~\ref{fig:continual}, the role of red weights is to be frozen, blue weights are randomly initialized and free to tune, and green weights are selectively initialized and consolidated to encourage forward transfer (Section~\ref{sec:forwardtransfer}). 
This method of expansion differs from Progressive Neural Networks \citep{rusu2016progressive} in that rather than connecting the green links to an adapter and \textit{adding} the adapter output to the input of the new layer, all inputs to the new layer are simply concatenated.
When learning each new classification task, minimization on the loss function (Equation~\ref{eqn:loss}) is applied only on that classifier with data for that task. 

An important question to ask when learning a new task, $T_{j}$, is that of how many new neurons need to be added. This is a difficult question, influenced by many factors, including how complex the new task $T_{j}$ is, and whether the previously learned tasks can help in learning $T_{j}$ (positive transfer; see Section~\ref{sec:forwardtransfer}). 
%
%
In Section~\ref{sec:exp_continual} we will propose a simple and effective strategy for the network expansion, and show that indeed, when learning a sequence of tasks with the consolidation values controlled as described here, previously learned tasks will not be affected.  

On the other hand, in Section~\ref{sec:exp_graceful}, we will show that if $\pmb{b}$ is decreased for some previous task, the performance on the task will degrade. In this case, graceful forgetting has happened.

\subsection{Forward Transfer}
\label{sec:forwardtransfer}


While continual learning without forgetting ensures past task performances are \textit{maintained}, previous tasks do not \textit{benefit} the learning of new tasks, a concept prominent in multi-task and transfer learning \citep{pan2009survey, zhang2017survey}, and appears in \LLL as ``forward transfer''.

\begin{figure}[ht]
    \centering
    \includegraphics[width=0.9\textwidth]{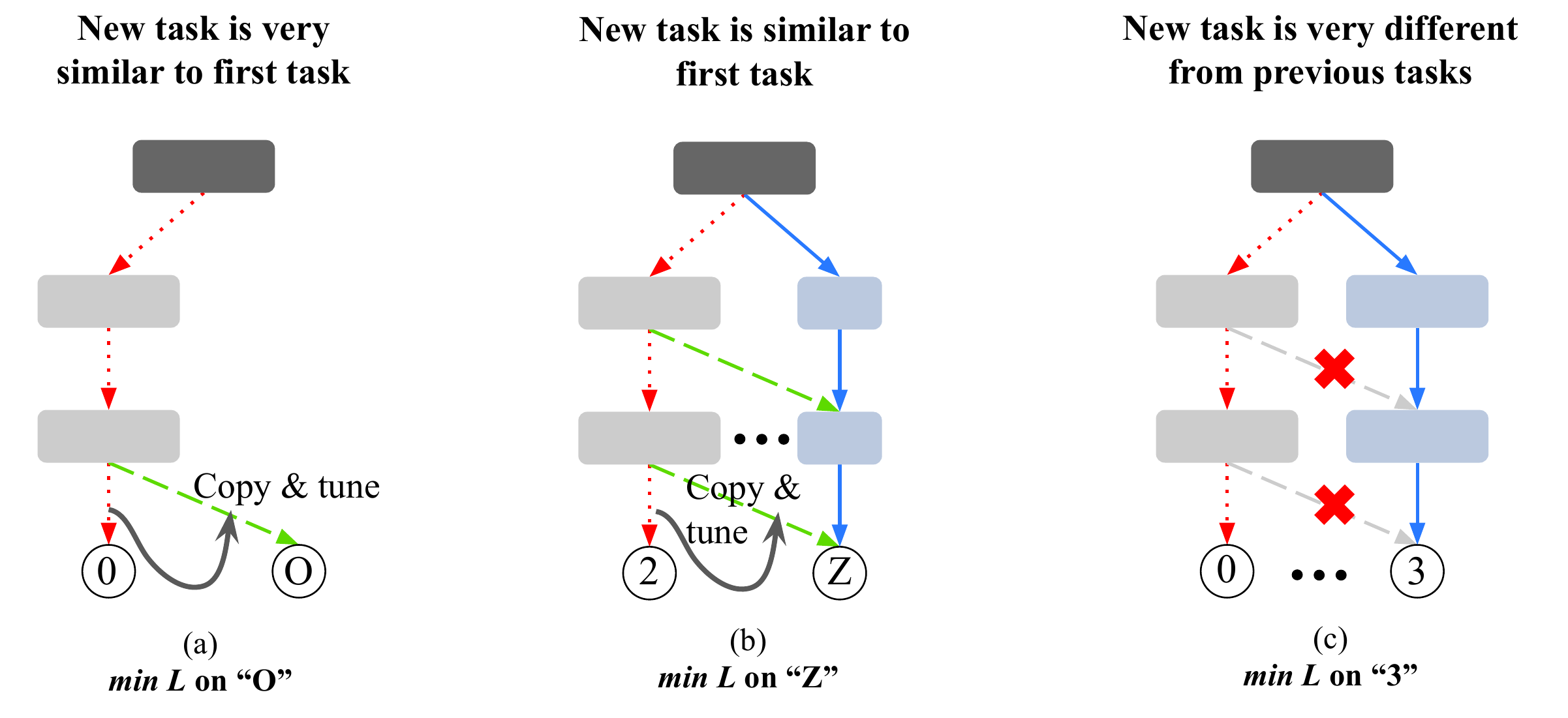}
    \caption{Adding proper forward transfer in continual learning without forgetting in our \LLL framework.  Three special cases of forward transfer are considered here.  In (a), when the new task (such as ``O'') is very similar to a previous task (such as ``0''), no new nodes may be needed (i.e., no network expansion as in Section~\ref{sec:continual}). The output-layer weights of the new task can be copied from the previous similar task during weight initialization. Note that those initial weights can still be fine-tuned in minimizing the loss of the new task (``O'' here).  In (b), if the new task (such as ``Z'') is similar enough (according to some threshold, see later) to a previous task (such as ``2''), positive transfer is expected from the previous task to the new task. Here, the expansion amount for the new task can be smaller, if the similarity is high.  All green and blue weights will be tuned when minimizing the loss of the new task (``Z'' here).  In (c), if the new task is very different from any previous tasks, the transfer links are disconnected to prevent possible negative transfer.  When positive transfer happens, we expect that the training requirements would be reduced significantly, i.e., few-shot learning would occur (see experiments in Section~\ref{sec:exp_forward}).} 
    \label{fig:forward_transfer}
\end{figure}

Forward transfer in deep neural networks is easily supported in our \LLL framework.  
Figure~\ref{fig:forward_transfer} illustrates several cases to consider with different levels of forward transfer.  
When a new task (such as ``O'') arrives, we first 
check if it is similar enough to a previous learned task (such as ``0'', ``1'', ...).  
Similarity can be estimated many different ways, such as a measure of the similarity of the data, how well the learned classifier performs on the new task, and so on. 
In Section~\ref{sec:exp_forward} we explain how we judge if two classification tasks are similar in more detail.
Then one or several previous similar tasks would be chosen to forward-transfer their learned knowledge to the new one.  Again, many possibilities exist.  Here we describe a simple strategy, reflected in Figure~\ref{fig:forward_transfer}. For the output layer weights of the new task, copy the weights of the output layer from the most similar previous task. And for the intermediate layers, only randomly initialize the green weights when the task similarity is above some threshold.

Our forward transfer mechanism is intended to have the effect that if a new task is very similar
to a previous task, positive transfer will occur, allowing the new task to be learned with less training data. This achieves few-shot learning. 
As we will show in Section~\ref{sec:exp_forward}, after learning ``0'', ``1'', ``2'', ``3'', and if ``O'' is to be learned next, since ``O'' is highly similar to ``0'', output weights of ``0'' would be copied to learn ``O''.  We will show that ``O'' is learned to a low testing error with many fewer examples. 
We also show that if forward transfer is incorrectly ``forced'' from dissimilar tasks when learning ``O'' and ``Z'', the predictive error is significantly higher.
By observing differences in predictive error across transfer strategies, we can distinguish between and quantify positive and negative transfer.

\subsection{Confusion Reduction}
\label{sec:confusionreduction}

As discussed in Section~\ref{sec:properties}, in our \LLL framework, tasks are learned in sequence, and data for the new tasks are only available when learning the new tasks.  
In this case, ``confusion'' can happen between two similar tasks 
(such as ``0'' and ``O'', ``2'' and ``Z''). 
It is interesting to note that the more similar the two tasks are, the more forward transfer would help in learning the new task (as seen in Section~\ref{sec:exp_forward}), yet at the same time, the more confusion occurs between them.  
In general, small confusion can happen between any pair or subset of classification tasks, and when the confusion is above some threshold, we need ways to reduce it\footnote{Note that we assume classes are mutually exclusive in this paper. In the case of multi-label learning, certain ``confusions'' may be desired, and need not be resolved. Also, when learning and evaluating each task individually, it is also called ``multi-head'' evaluation. 
When evaluation is done across all classifiers learned so far, it is often called ``single-head'' evaluation  \citep{chaudhry2018riemann}. 
Single-head evaluation is much harder, thus in our \LLL framework, confusion between classifiers must be reduced.}.

We propose to resolve such confusion in a pairwise manner. 
To resolve confusion between $T_{i}$ (such as ``0'') and $T_{j}$ (such as ``O''), some training data of both $T_{i}$ and $T_{j}$ would be needed, whether stored or generated.
Here, confusion is measured by the sum of errors evaluated on $T_{i}$ and $T_{j}$ when data of both tasks is presented as input.  The training is to minimize the error on these tasks together, while weights for other tasks are frozen.   

\begin{figure}[ht]
    \centering
    \includegraphics[width=0.85\textwidth]{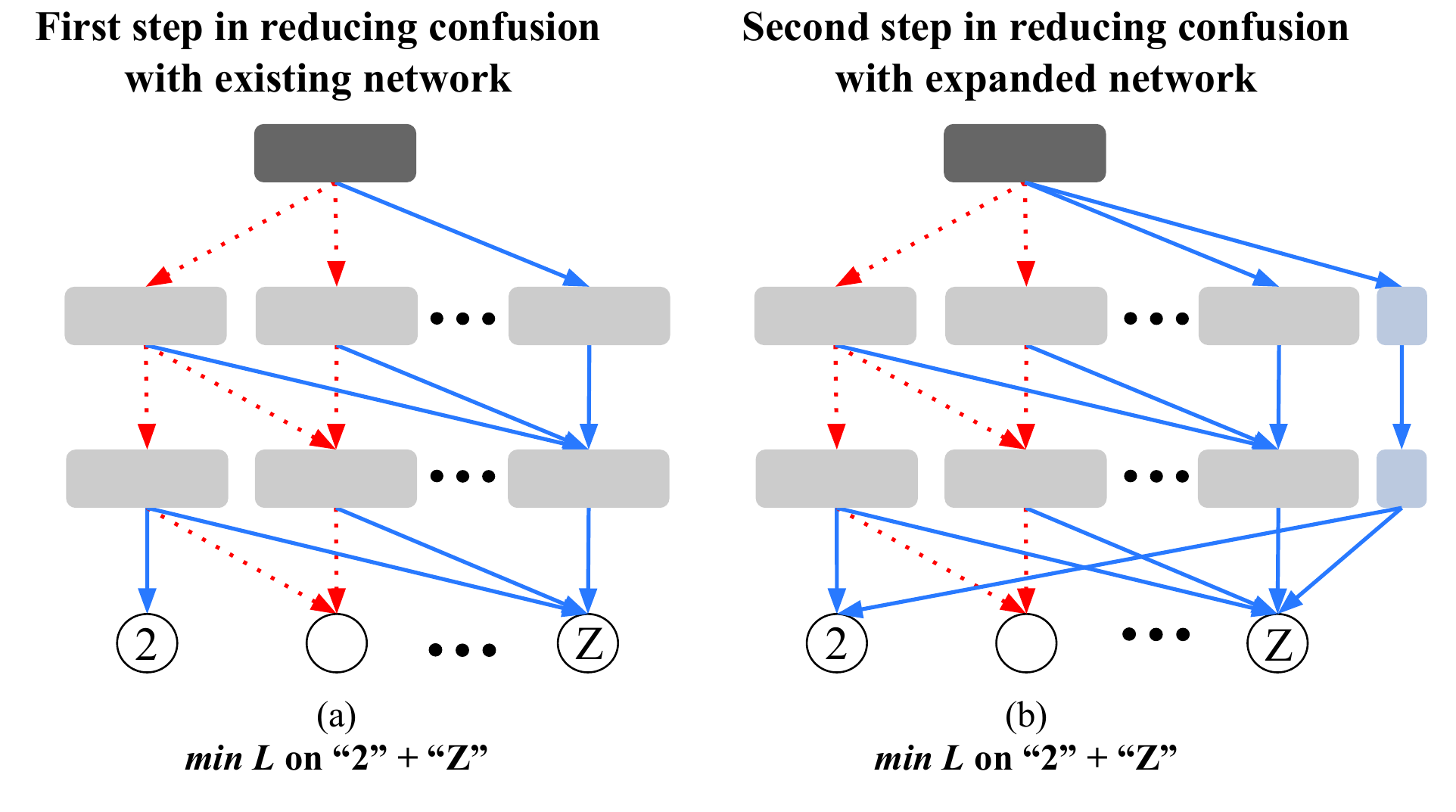}
    \caption{The two steps of the confusion-reduction process. In (a) we first try to reduce confusion using the existing network capacity, freezing the necessary weights to maintain performance on other tasks. In (b), where confusion cannot be reduce by fine-tuning alone, we expand the network and all newly added weights can be learned.}
    \label{fig:confusion}
\end{figure}

More specifically, 
whenever confusion between $T_{i}$ and $T_{j}$ is greater than some threshold, $\gamma \in [0, 1]$, we propose a two-step strategy to reduce confusion.
First, using the existing network, simultaneously fine-tune the last-layer weights of $T_{i}$ and weights of $T_{j}$ on samples of the confused tasks. 
This step is shown in Figure~\ref{fig:confusion}a.
When the confusion is small, this step alone may reduce the confusion to below $\gamma$.
If the current network capacity is not sufficient to resolve confusion, then we can move onto the second step, where we expand the model by some amount, and all the new weights can now be learned. 
This step is reflected in Figure~\ref{fig:confusion}b.
In both cases, only those weights associated with the confused tasks are tuned, leaving other tasks unaffected.

In experiments described in Section~\ref{sec:exp_nonconfusion}, we demonstrate that both steps of this process contribute to decreasing confusion between highly similar tasks.

\subsection{Graceful Forgetting}
\label{sec:gracefulforgetting}

As we learn more and more tasks in our \LLL framework, the expanding network may reach some size limitation.  Eventually, we may reach a limit where the network cannot offer free units to learn a new task well without significantly forgetting already-learned tasks. 
In such cases, we can consider ``graceful'' forgetting \citep{aljundi2018memory} of previous tasks.  

\begin{figure}[ht]
    \centering
    \includegraphics[width=0.85\textwidth]{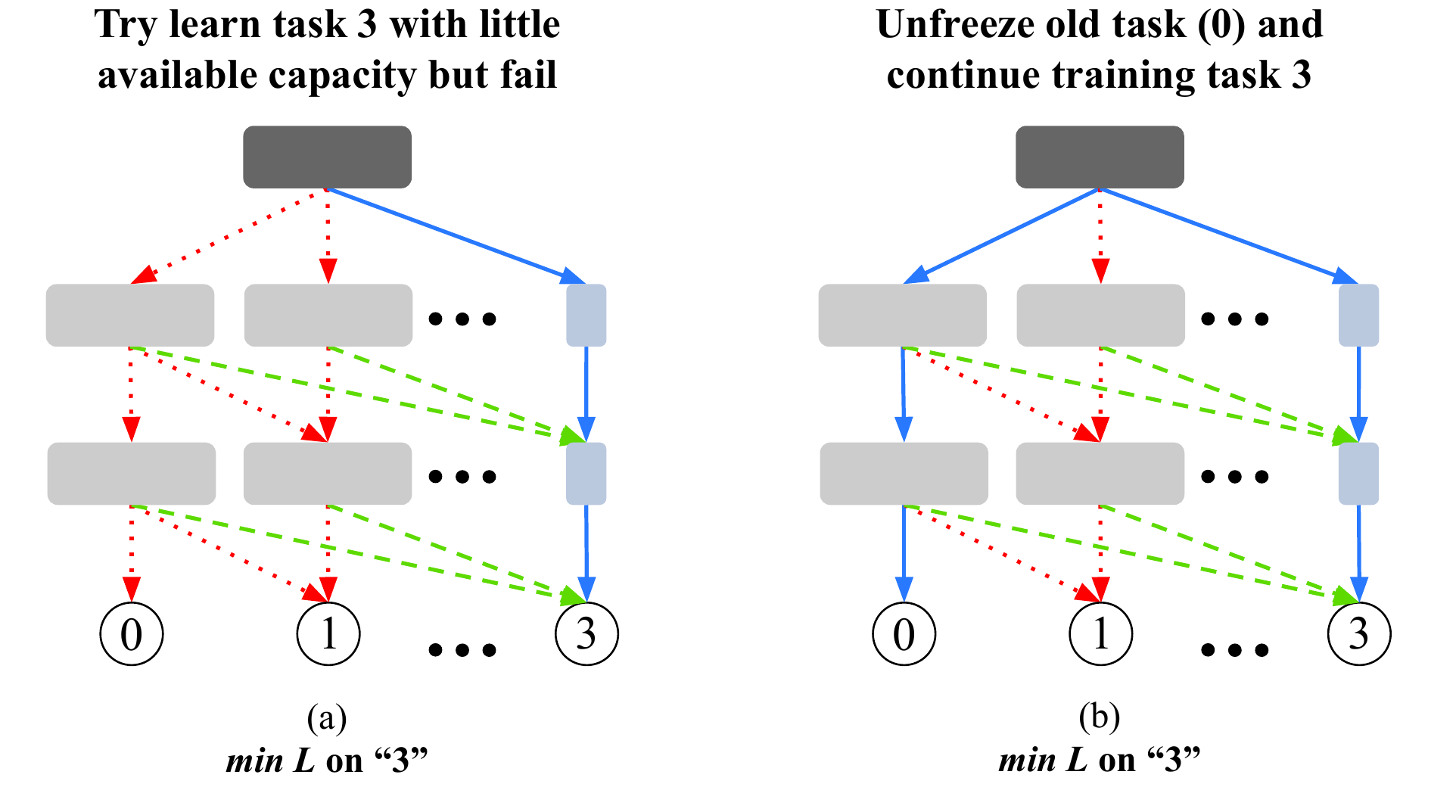}
    \caption{Graceful forgetting of an old task when learning a new task without free network capacity.  In (a) when minimizing the loss on ``3'', it cannot be reduced further. In (b), task ``0'' is chosen to be forgotten, and its weights (blue links) are unfrozen ($\pmb{b}$ is set small or 0).  Training on task ``3'' continues to achieve a much smaller loss as previous tasks are gracefully forgotten.}
    \label{fig:graceful_forgetting}
\end{figure}

For example, consider the case where after learning $T_{0}, T_{1}, T_{2}$, there are very few free units to learn $T_{3}$. 
If we freeze all previous task weights with $\pmb{b} = \infty$ and train on data of $T_{3}$, we will find that $T_{3}$ cannot be learned well even after many epochs of training (see Figure~\ref{fig:graceful_forgetting}a and Section~\ref{sec:exp_graceful}). In this case, we can set the $\pmb{b}$ values for some less important tasks, say $T_{0}$, to be small or 0, and then train with samples of $T_{3}$ again, as in Figure~\ref{fig:graceful_forgetting}b. 
In this case we expect to see that the error on $T_{3}$ decreases, while sacrificing performance of $T_{0}$ gradually and gracefully. As $T_{0}$ also indirectly affects $T_{1}$ and $T_{2}$ (even though their $\pmb{b} = \infty$), their predictive accuracy may also be slightly decreased.  

One might wonder why $T_{0}$ would be chosen to be gracefully forgotten. 
This choice is likely to be domain and situation specific.  For example, if the $T_{0}$ classifier has not been needed to make predictions for a long period of time, its ``importance'' may be lower than other classifiers, and its $\pmb{b}$ values could be reduced gradually to allow graceful forgetting.  This would be especially useful when there are no more free neurons for learning new tasks.  In Section~\ref{sec:parallels}, we will discuss parallels with human leaning, which seems to exhibit a similar phenomenon.  For example, a friend who you have not met and thought about for a long time may tend be to forgotten gradually over years. 

\subsection{Backward Transfer}
\label{sec:backwardtransfer}

\begin{figure}[ht]
    \centering
    \includegraphics[width=0.4\textwidth]{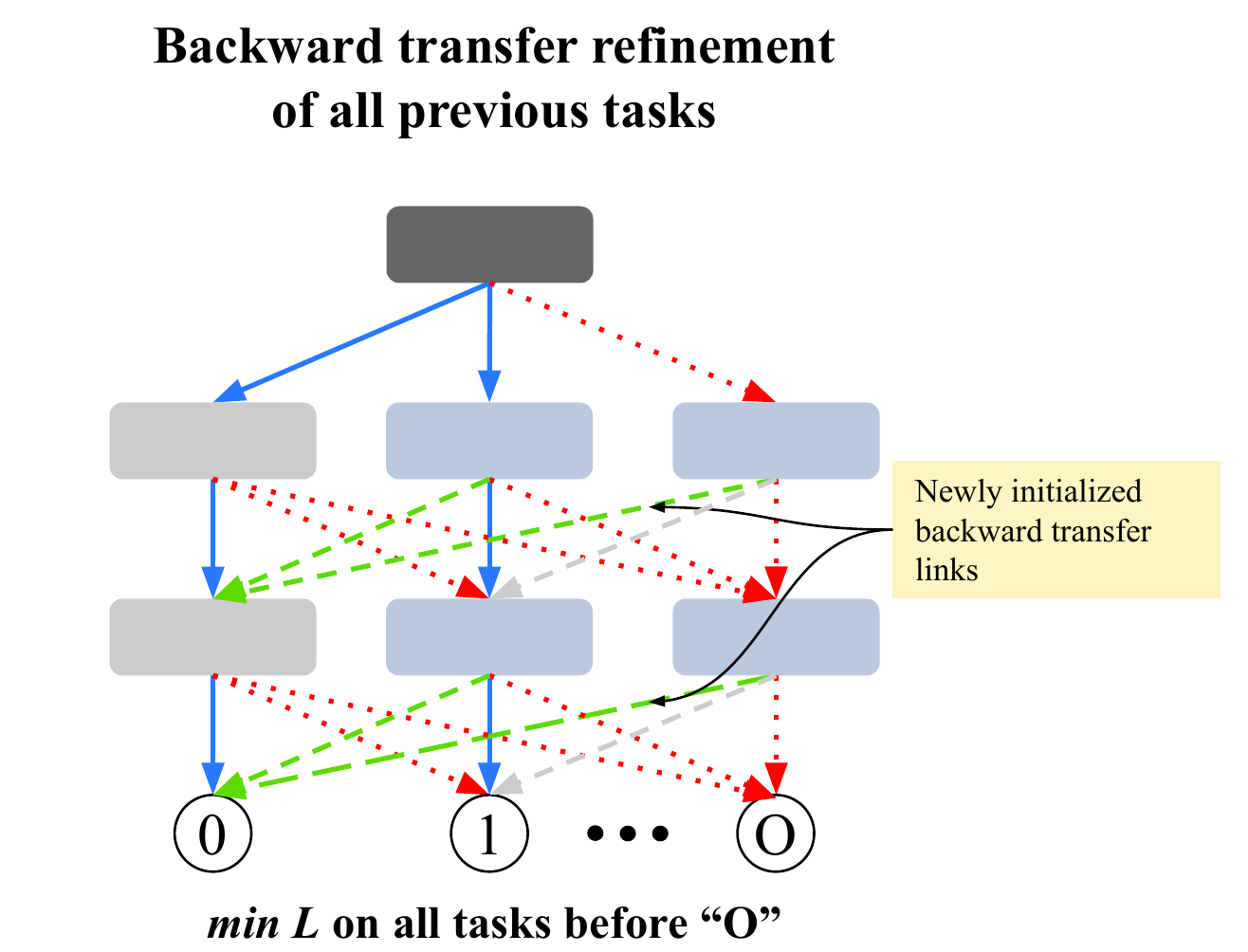}
    \caption{Consolidation of weights during backward transfer from ``O'' to previous tasks (``0'', ``1'', ...).
    Since the earlier task has already been trained, this process would require little training time to converge (as only backward links are learned from scratch).}
    \label{fig:backward_transfer}
\end{figure}

In Section~\ref{sec:forwardtransfer} we discussed forward transfer, where previous tasks help to learn the new task. Can new tasks similarly help to improve performance on previous tasks to achieve positive backward transfer?
Our framework may be able to achieve such backward transfer by initializing backward transfer links (between sufficiently similar tasks) and fine-tuning on the desired tasks. In Figure~\ref{fig:backward_transfer}, the links supporting backward transfer from ``O'' to ``0'' are labelled. 
In Section~\ref{sec:exp_backward} we evaluate this approach to backward transfer.  Similar to forward transfer, we find that when tasks are sufficiently similar, backward transfer works well. This appears similar to human learning, where learning related concepts reinforce each other.

Although forward and backward transfer, as well as confusion reduction are performed with pairs of tasks, it is possible to extend it to operate on larger subsets of tasks, or even all tasks at once, to perform an ``overall refinement''. 
%
%
Such overall refinement is obviously similar to the more resource-intensive batch multi-task learning setting, where we assume that we can train on all tasks at once. However, in our \LLL setting, after already taking care to achieve forward transfer and confusion reduction separately and more efficiently, 
this refinement process may be done infrequently and quickly.   

\section{Experimental Verification}
\label{sec:experiments}


In this section we will present  
experimental results demonstrating the capabilities of our proposed \LLL framework.
These experiments mainly use the running example in Figure~\ref{fig:task_seq}, and thus we consider them as proof-of-concept experiments.  

\paragraph{Task sequence} We will use the binary classification task sequence illustrated in Figure~\ref{fig:task_seq}, with samples taken from the balanced EMNIST dataset~\citep{journals/corr/EMNIST}.
This task sequence is a minimal case allowing for proof-of-concept experiments where we can be sure that there is a) clear room for forward transfer (e.g. from ``0'' to ``O'' or ``2'' to ``Z'') and b) clear cases of confusion (e.g. between ``0'' and ``O''). In a more complex task sequence it would be difficult to verify whether the proposed mechanisms work as intended.
All results will be averaged across 15 random seeds.
After first establishing feasibility with these toy experiments, it is our future work to verify the proposed \LLL algorithms on more complex task sequences.  
In addition to using simplified experiments, we only set $\pmb{b}$ values to $\infty$ or $0$ in our experiments, equivalent to masking selected weights during gradient descent. It would be our future work to set $\pmb{b}$ to intermediate values in order to observe more graded behaviors of our \LLL algorithms.
    
\paragraph{Architecture and training} We will use a network architecture with two hidden layers with ReLU activation. The Adam optimizer \citep{kingma2014adam} will be used, with the default hyperparameters provided by Keras \citep{chollet2015keras}. We will use a batch size of 64 and training for 10 epochs for each task, unless otherwise stated. Further details will be given for each experiment.

The outline of this section is as follows. 
In Section~\ref{sec:exp_continual}, we will evaluate our framework on continual learning without forgetting. 
In Section~\ref{sec:exp_forward}, we will evaluate it at accelerating task learning with forward transfer. 
In Section~\ref{sec:exp_nonconfusion}, we will evaluate its ability to reduce confusion between tasks. 
In Section~\ref{sec:exp_graceful}, we will evaluate its ability to gracefully forget tasks to allow for new-task learning.
Finally, in Section~\ref{sec:exp_backward}, we will evaluate its ability to support backward transfer.


\subsection{Continual Learning of New Tasks Without Forgetting}
\label{sec:exp_continual}

To evaluate the ability of our proposed framework to continually learn new tasks without forgetting, we will observe the test AUC (area under the receiver operating characteristic) of each task as more tasks are learned. AUC is a suitable metric for evaluating the performance on these binary tasks as a result of the imbalanced data (for each task, the ratio of positive to negative samples is 1:4). When the AUC on a task remains constant after it is initially learned, it indicates that forgetting has been avoided. For this experiment, we will use a constant network expansion rate of 25 units per hidden layer per task. All forward transfer links (green links in Figure~\ref{fig:continual} will be enabled and randomly initialized. Additionally, we will use 100 positive samples for each task (and 400 negatives -- 100 per character type in the negative class).

\begin{figure}[ht]
    \centering
    \includegraphics[width=0.75\textwidth]{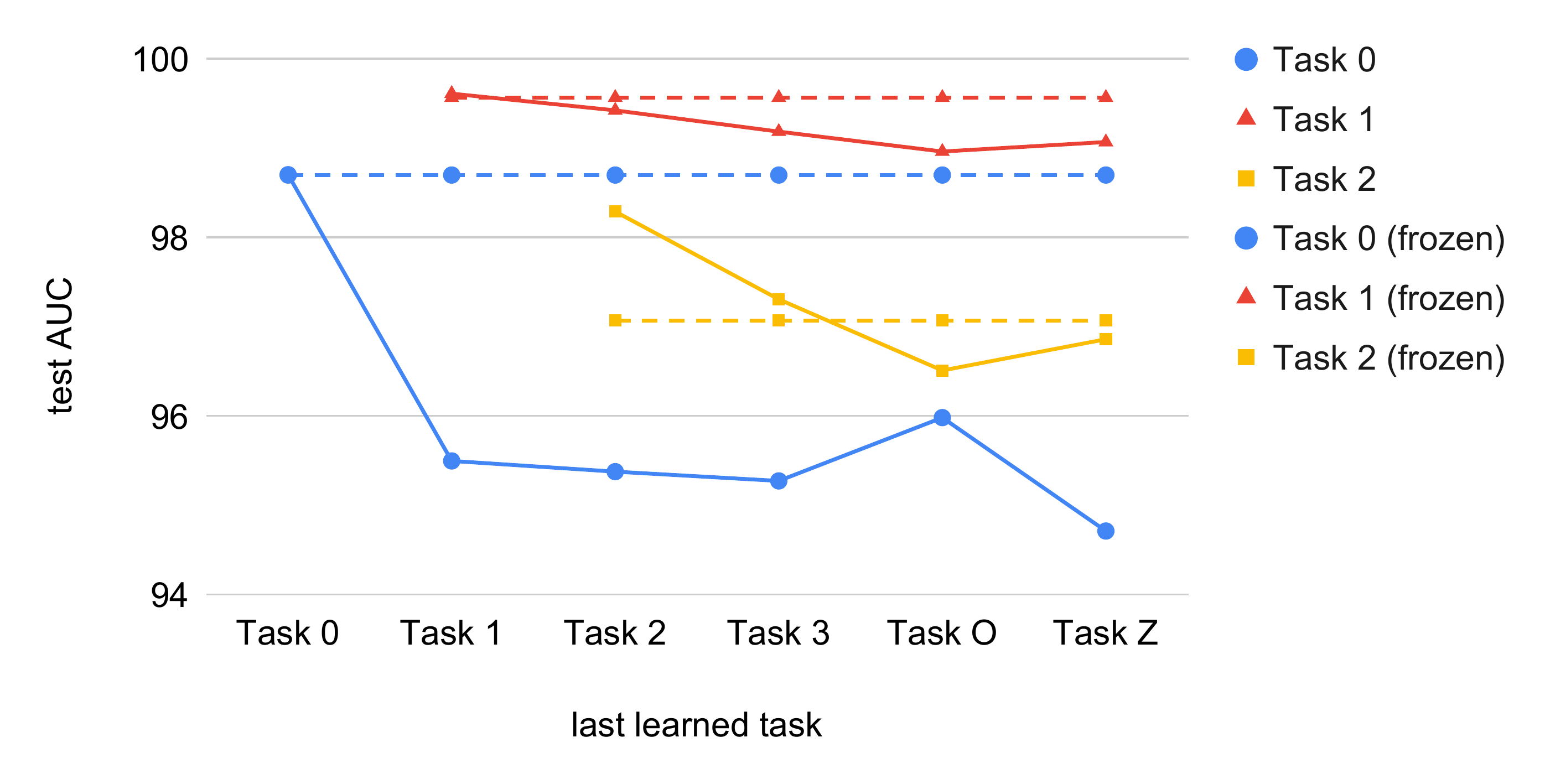}
    \caption{Evaluating the non-forgetting ability of the proposed framework. Solid lines indicate the performance histories of each task when no freezing is used. Dashed lines correspond to performance with freezing. To avoid clutter, only the histories of the first three tasks are shown.}
    \label{fig:continual_results}
\end{figure}

\paragraph{Observations} From Figure~\ref{fig:continual_results}, we can clearly see that enabling the non-forgetting mechanism by freezing learned task weights indeed prevents forgetting, as indicated by the dashed lines remaining at a constant AUC. While the initial performance of later tasks might be lower than without freezing, eventually freezing pays off. For example, while the learned task ``2'' performance is lower when non-forgetting is enabled (because there are fewer tunable weights in the network), eventually the performance of the unfrozen model drops even lower (once task ``O'' is learned). 

\subsection{Forward Transfer}
\label{sec:exp_forward}

To evaluate the ability of our framework to support forward transfer, we will observe the test AUC of tasks using a variety of initialization and weight copying strategies. 
For this experiment, we will enable non-forgetting and also use difficulty-based expansion (up to a maximum expansion rate of 25 per layer per task). 

\paragraph{Similarity and difficulty-based expansion method} To accommodate a new task, $T_{j}$, we extend the neural network width by an amount, $N_{j}$, proportional to the estimated difficulty of the task. 
To compute $N_{j}$, we first compute the maximum similarity to previous tasks. To compute the similarity between two tasks, $sim(T_{i}, T_{j})$, we feed positive samples of the new task, $T_{j}$, into the existing network, and average the probabilities output by model for $T_{i}$. 
When the similarity between $T_{j}$ and any previous task is high (i.e. the new samples are similar to those of a previous task), proportionally fewer nodes are added. That is, $N_{j} = N_{max} \left(1 - \max_{i = 1, ... j-1}sim(T_{i}, T_{j})\right)$. In these experiments, $N_{max} = 25$.
In the extreme case where a new task is identical (or very similar) to a previous one, no new nodes (aside from the output) may need to be added.

For the first four tasks, we will use 100 positive samples for each task, but only 10 for the last two tasks, so that the differences between the various forward transfer strategies can be highlighted. The four strategies we compare are as follows:


\begin{itemize}
    \item \textsc{AllRandomInit}: This strategy simply randomly initializes all forward transfer links (as in the previous experiment).
    
    \item \textsc{OneSimilar}: This strategy first computes the similarity between all previous tasks and the new one. When the similarity is above $\alpha = 0.5$, then the corresponding forward transfer links in the intermediate layers are randomly initialized. Additionally, when the single most similar previous task has a similarity $ > \alpha$, then the output layer weights are copied from that task, as reflected in Figure~\ref{fig:forward_transfer}.
    
    \item \textsc{OneRandom}: This strategy is similar to the previous one, except that one random previous task is chosen to copy weights from, independent of the similarity.
    
    \item \textsc{OneWorst}: This strategy initializes only the intermediate layer forward transfer links from the least similar task, and copies output weights only from the least similar task.
\end{itemize}

\begin{figure}[ht]
    \centering
    \includegraphics[width=0.75\textwidth]{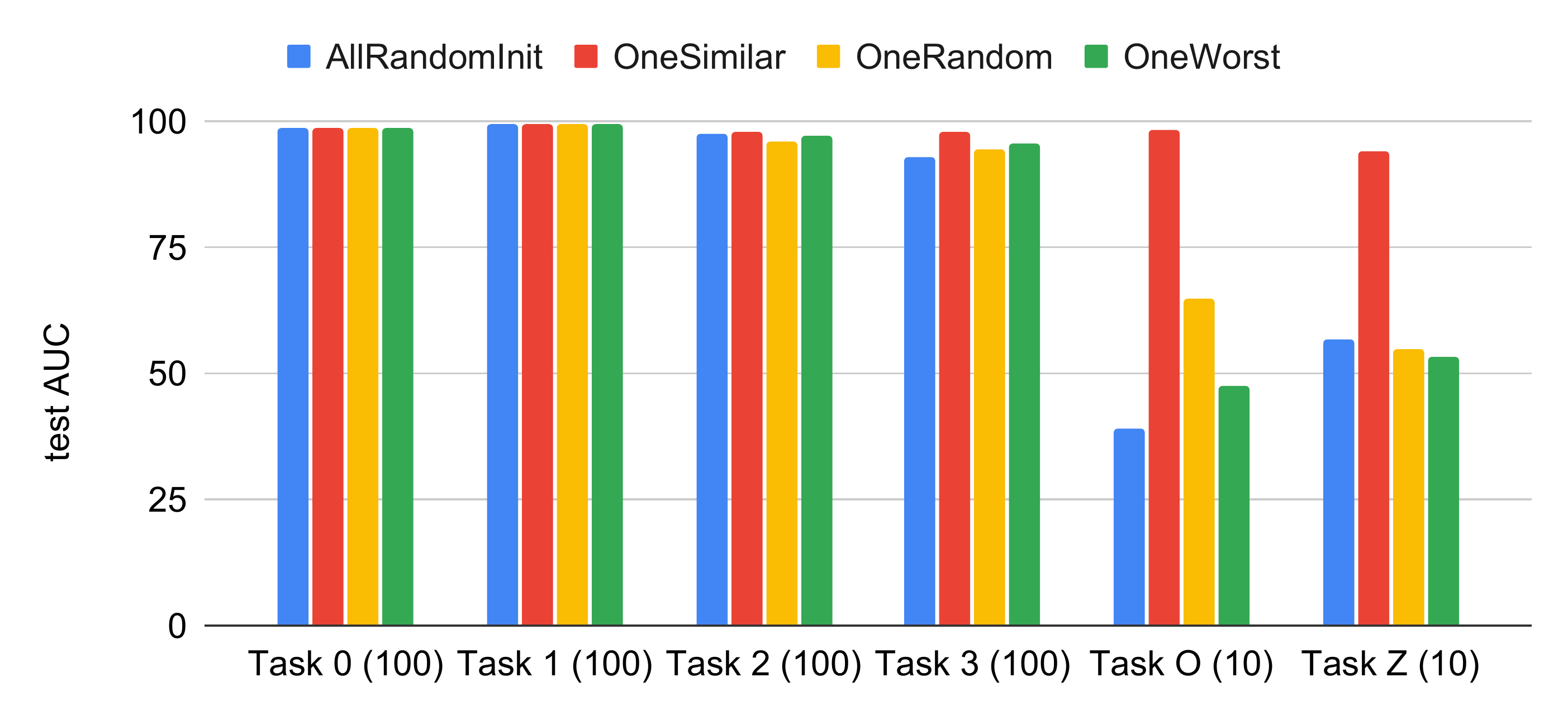}
    \caption{Evaluating the proposed forward transfer mechanism of our framework (shown here as \textsc{OneSimilar}). The four strategies shown in the figure are described in the text. The first four tasks have 100 positive samples each, while the last two have only 10, requiring few-shot learning. 
    Our proposed weight initialization and copying mechanism (red), performs the best.  It allows for achieving much higher test AUCs than alternative strategies.}
    \label{fig:forward_transfer_results}
\end{figure}

\paragraph{Observations} From Figure~\ref{fig:forward_transfer_results}, we can see that on the last two few-shot tasks, the \textsc{OneSimilar} strategy works best by far, achieving comparable results to tasks with 10X as much data. This is of course possible via the clear similarity between ``O'' and ``0'' and between ``Z'' and ``2''. The \textsc{OneRandom} strategy, which works similar to the \textsc{OneSimilar} strategy for the intermediate layers, performs second best, suggesting that while the weight copying from the most similar task has the largest benefit, initializing transfer links from similar tasks is also helpful.

\begin{figure}[ht]
    \centering
    \includegraphics[width=1\textwidth]{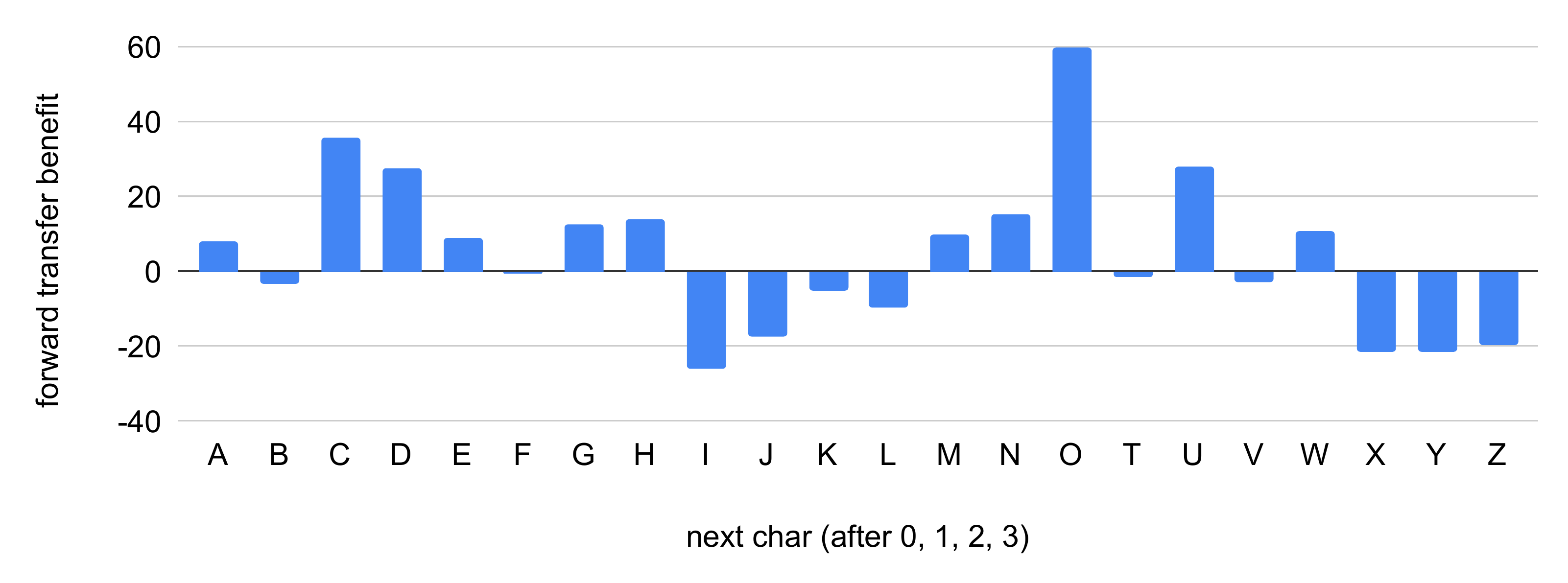}
    \caption{The test AUC difference between using the \textsc{OneAlways} and \textsc{AllRandomInit} strategies for a range of characters. Positive values indicate that \textsc{OneAlways} performs better. The characters here are learned after ``0'', ``1'', ``2'', ``3''. Note that ``P'', ``Q'', ``R'', and ``S'' are skipped, as they are included in the negative class.}
    \label{fig:forward_transfer_benefit_results}
\end{figure}

In Figure~\ref{fig:forward_transfer_benefit_results}, we can see what happens when, instead of only copying weights when a previous task is similar enough (i.e. \textsc{OneSimilar}), we ``force'' weights to be copied even if the most similar task is not very similar. We call this strategy \textbf{\textsc{OneAlways}}. After first learning ``0'', ``1'', ``2'', and ``3'' with 100 positive samples, we see that for many options for the fifth task (again with only 10 positive samples), the achieved performance is better than the baseline strategy of \textsc{AllRandomInit}. There are however, several characters where this strategy hurts performance (e.g. ``I'', ``J'', ``X''), which provide clear cases of negative transfer. 

\subsection{Confusion Reduction}
\label{sec:exp_nonconfusion}

This experiment will evaluate the ability of our framework to reduce confusion between tasks, using the two-stage $\pmb{b}$-setting and expansion strategy illustrated in Figure~\ref{fig:confusion}. We will observe the maximum confusion of the last two tasks (``O'' and ``Z'') through the stages of confusion reduction -- lower values indicate better confusion reduction. Confusion between a pair of tasks is the percentage of the time that positive samples from either task are mis-classified as the other task. For ``O'', the maximum confusion almost exclusively refers to confusion with ``0'', and for ``Z'', it is with ``2''.

For this experiment, we will now use the \textsc{OneSimilar} forward transfer strategy with 100 positive samples per task.
Additionally, we will use a confusion threshold of $\gamma = 0.1$, so that each stage of confusion reduction will run if the confusion is not yet below 10\%.
We will also evaluate confusion expansion amounts (second stage) of both 5 and 10 per layer.

\begin{figure}[ht]
    \centering
    \includegraphics[width=0.75\textwidth]{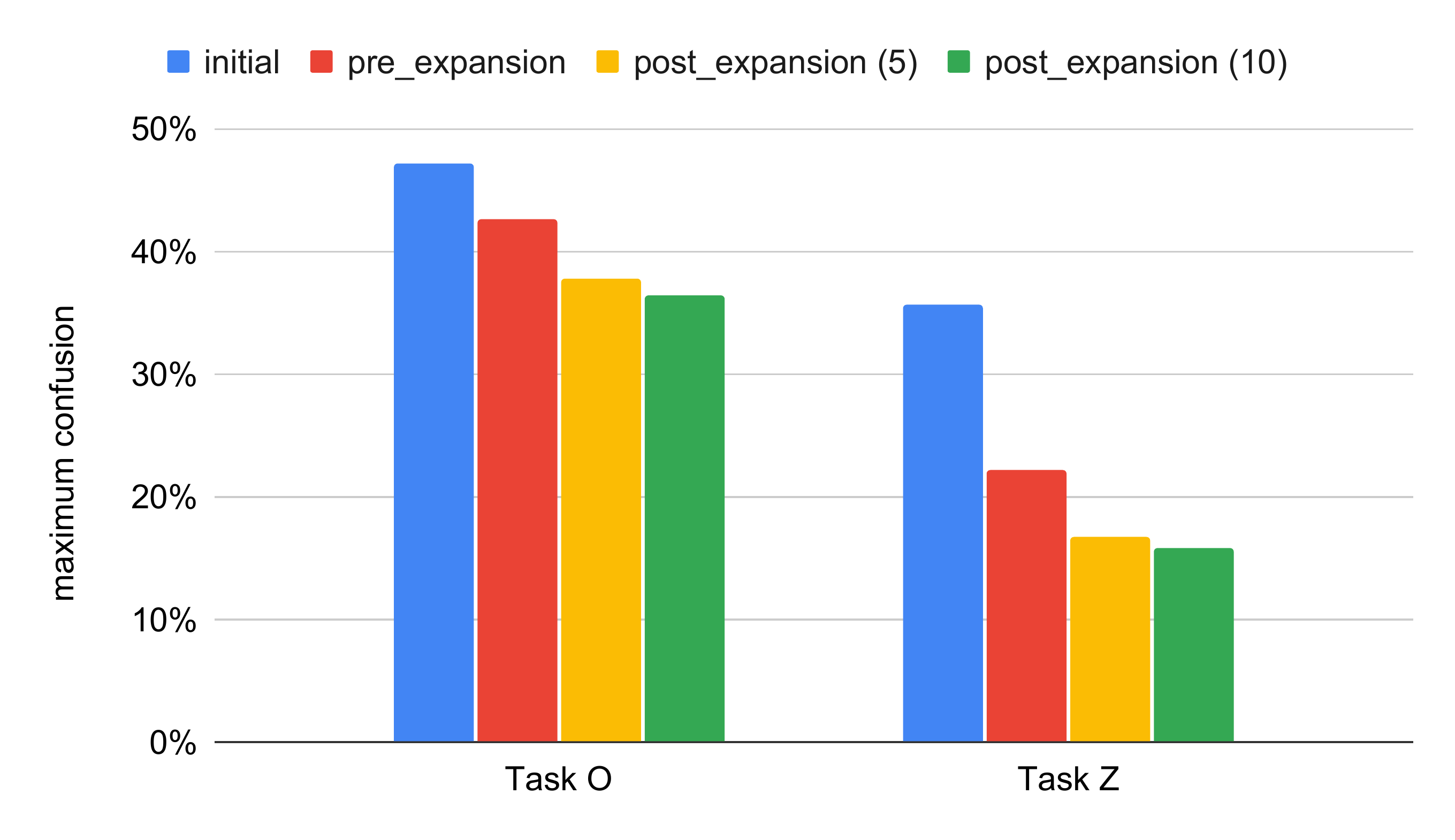}
    \caption{Evaluating the confusion-reduction effectiveness of the proposed framework. We report here the maximum confusion between ``O'' and previous tasks (``0'', ``1'', ``2'', ``3'') and between ``Z'' and previous tasks. Most often, confusion happens between ``O'' and ``0'' and between ``Z'' and ``2''.}
    \label{fig:confusion_results}
\end{figure}

\paragraph{Observations} From Figure~\ref{fig:confusion_results}, we can see that both stages of the confusion reduction mechanism (first tuning, i.e. pre-expansion and then last-resort expansion with tuning, i.e. post-expansion) contribute to reducing confusion. We also see that using a larger confusion-expansion amount contributes to further reduction of confusion.

\subsection{Graceful Forgetting}
\label{sec:exp_graceful}

To evaluate the ability of our framework to perform graceful forgetting, we will try to learn a new task, ``3'', with a very small number of new unfrozen nodes (after first learning ``0'', ``1'', ``2''), and observe the test AUC of all tasks over the course of training this task. Successful graceful forgetting should result in the performance of the new task substantially increasing while previous tasks experience low rates of forgetting. 

For this experiment, we will return to using a constant expansion of 25 nodes per task per layer, except for task 3, where only one new node per layer will be added. We will also enable non-forgetting (except where graceful forgetting is performed), and randomly initialize all forward transfer links. We will train task ``3'' for 10 epochs before performing the graceful forgetting, at which point we train it for another 10 epochs.

\begin{figure}
\begin{floatrow}
\ffigbox{%
  \includegraphics[width=0.45\textwidth]{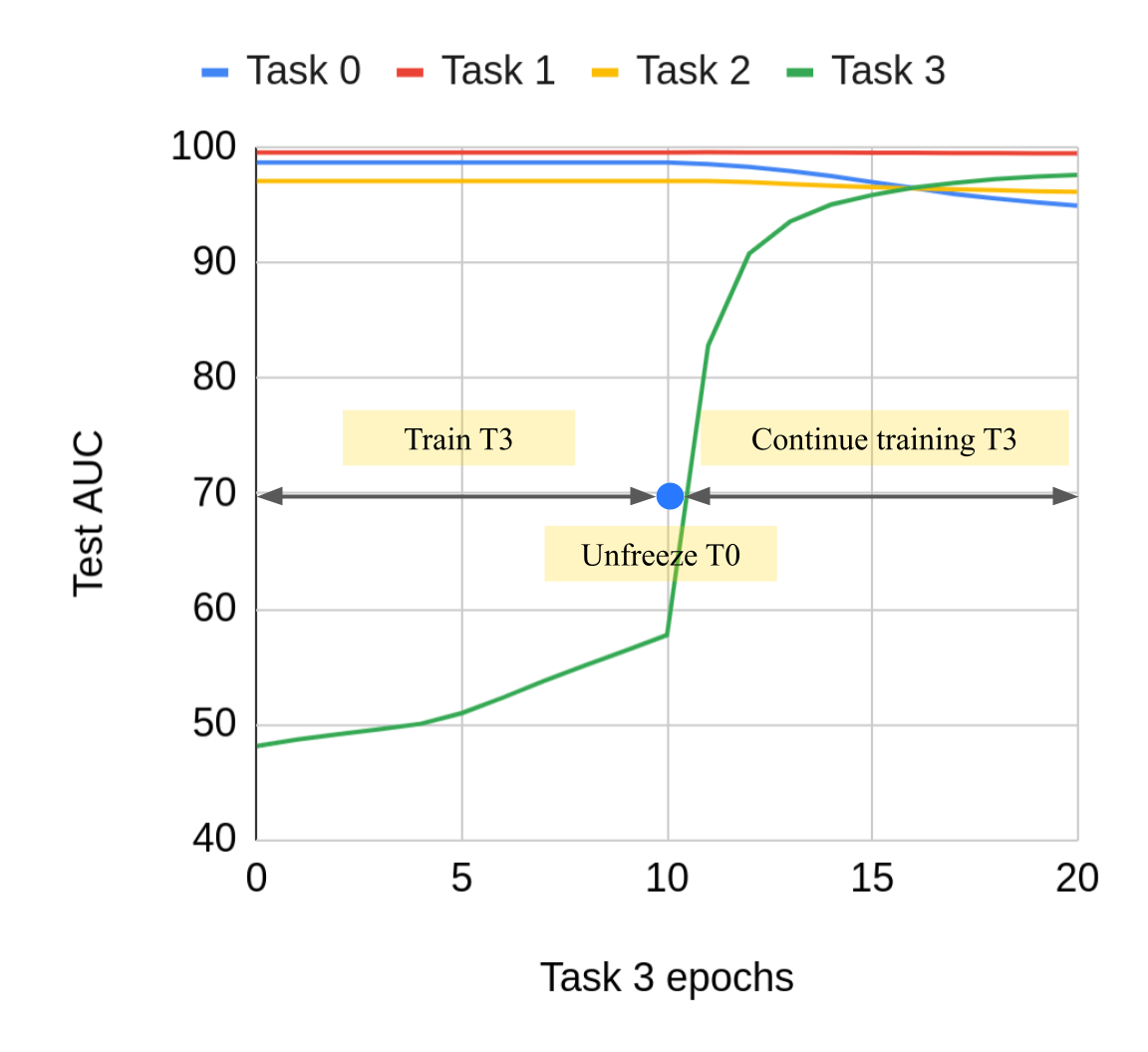}
}{%
  \caption{Graceful forgetting with forgetting of the first task at the 10 epoch mark. We find that performance on task ``3'' quickly increases after task 0 is unfrozen, with most of the forgetting occurring in task 0.}
  \label{fig:graceful_results_a}
}
\ffigbox{%
  \includegraphics[width=0.45\textwidth]{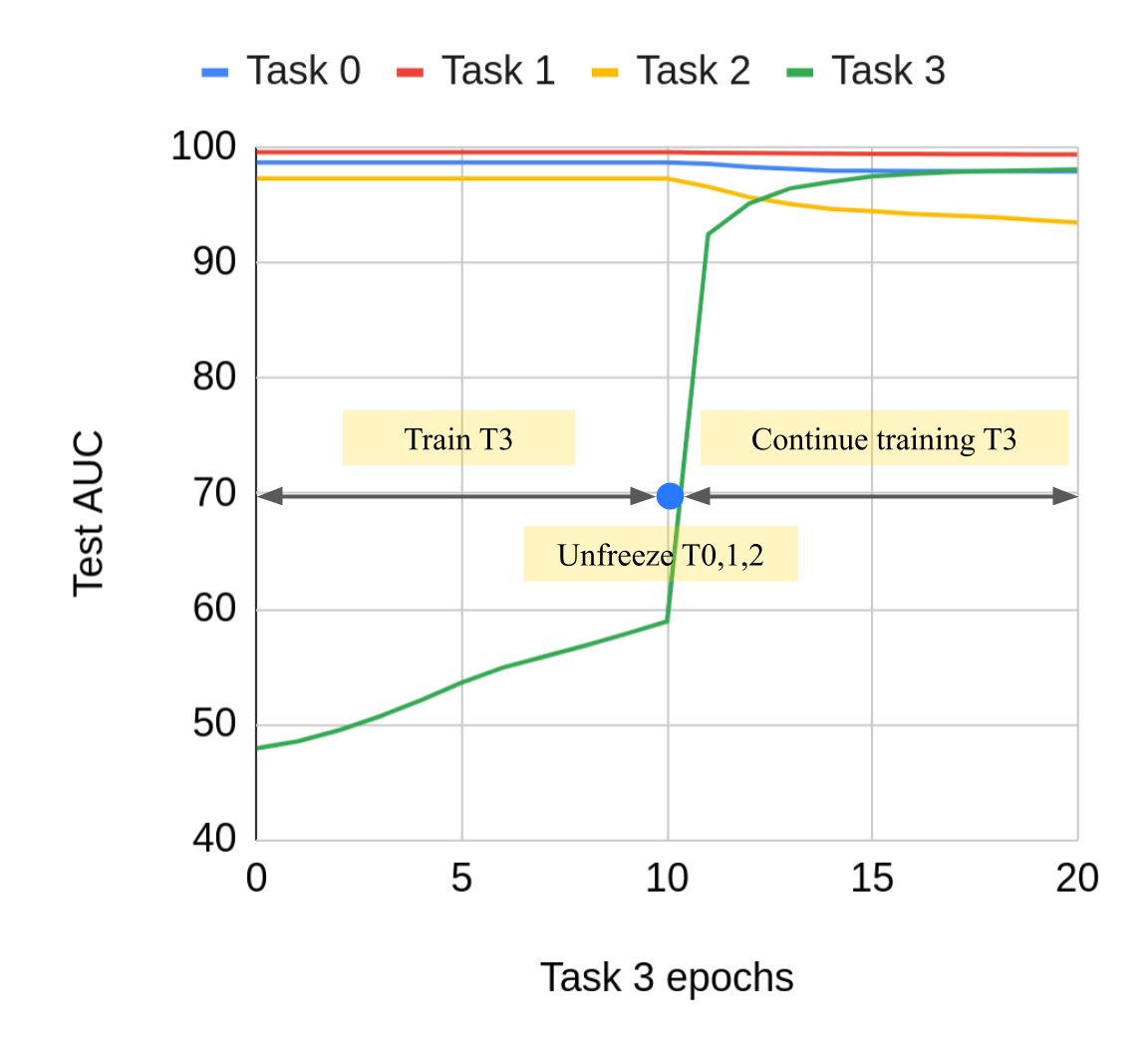}
}{%
  \caption{Graceful forgetting with forgetting of the first three tasks at the 10 epoch mark. We find that the learning of task ``3'' is faster following unfreezing of previous tasks, with more distributed forgetting.}
  \label{fig:graceful_results_b}
}
\end{floatrow}
\end{figure}

\paragraph{Observations} From Figures~\ref{fig:graceful_results_a} and \ref{fig:graceful_results_b}, we can see that graceful forgetting has a large positive effect on the learning of the fourth task, where it is only task ``0'' being forgotten (as in Figure~\ref{fig:graceful_results_a}) or ``0'', ``1'' and ``2'' being forgotten (as in Figure~\ref{fig:graceful_results_b}). When all three previous tasks are unfrozen, task ``3'' appears to improve faster as we may expect. In both cases, the forgetting experienced by the previous tasks is slow.

\subsection{Backward Transfer}
\label{sec:exp_backward}

This experiment will evaluate the ability of our framework to achieve backward transfer, where knowledge from newer tasks is used to allow older tasks to be learned better. This is done by initializing backward transfer links and performing fine-tuning of tasks. 

To evaluate the ability of our framework to achieve backward transfer of knowledge, we will first learn a task ``0'' with few samples, and learn
either a second similar task (``O'') or a second dissimilar task (``Z''). After learning the second task, we will initialize the backward transfer links (labelled in Figure~\ref{fig:backward_transfer}) and fine-tune ``0''. For this experiment we found it sufficient to set $\pmb{b} = 0$ for only the weights (both new and old) in the output layer for ``0''. We will observe the test AUC of task ``0'' over the course of fine-tuning. Effective backward transfer would result in the performance of task ``0'' increasing faster during tuning when the second task contains relevant knowledge (``O'').

For this experiment, we will use a constant expansion of 25 nodes per task per layer, with non-forgetting and random initialization of all forward transfer links (and backward transfer links when applicable). We will train the first task with only 10 positive samples and the second task with 50. Fewer samples for the first task will allow differences between backward transfer performance for different task sequences to be emphasized.

\begin{figure}[ht]
    \centering
    \includegraphics[width=0.85\textwidth]{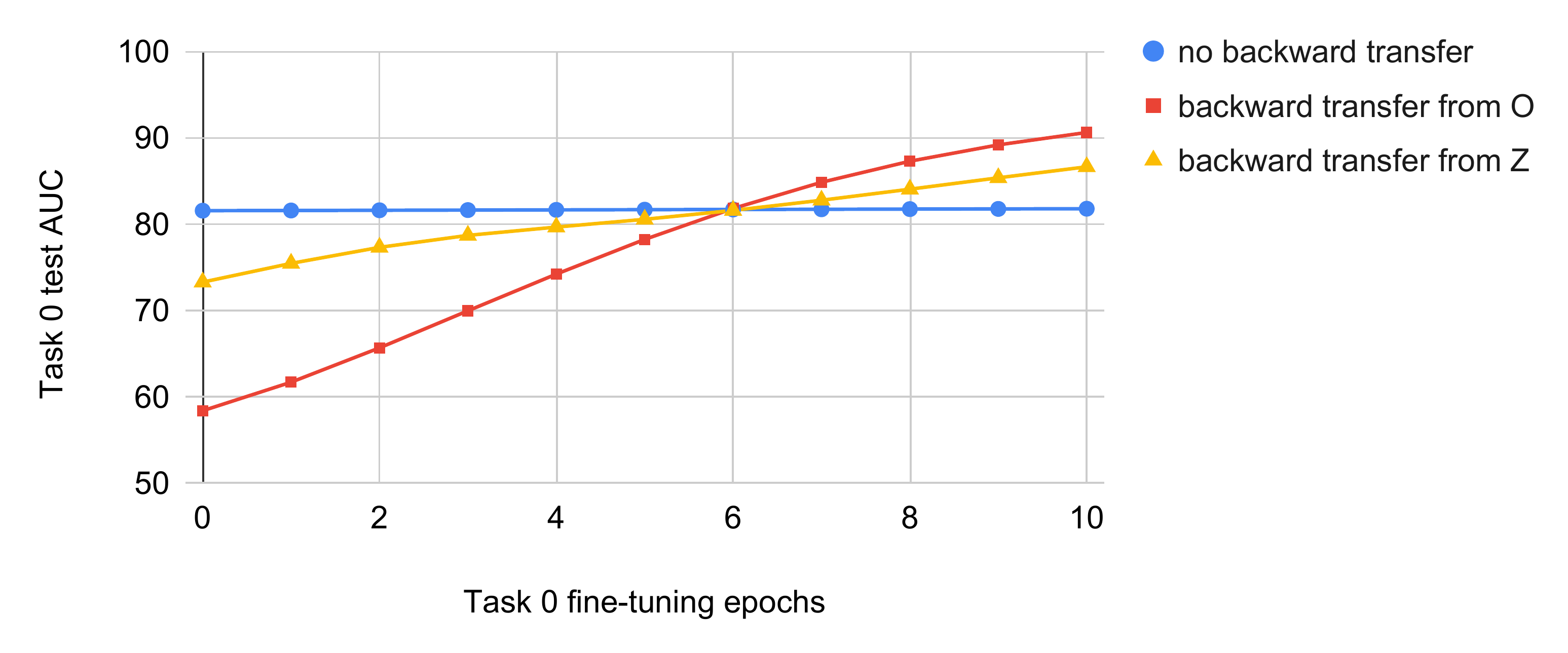}
    \caption{Exploring the backward transfer ability of the framework. We see that when the second task is similar to the first (``O'' is more similar to ``0'' than ``Z''), the benefit from tuning with backward transfer links enabled is greater.}
    \label{fig:backward_transfer_results}
\end{figure}

\paragraph{Observations} From Figure~\ref{fig:backward_transfer_results}, we can see that when no backward transfer links are enabled (blue line), there is almost no performance increase during tuning. This is expected since all weights being tuned have already been trained on the data for ``0''. In the cases where backward transfer links are enabled (only for the output layer in our experiments),
we see that the performance does drop at first. This is a result of the randomly initialized weights negatively influencing the classifier.
Once fine-tuning begins however, when the transfer links are from a similar task (``O'' -- red line), the performance of ``0'' increases to an overall higher level than when the transfer is from a dissimilar task (``Z'' -- yellow line). 
The performance of ``0'' does increase in both cases however, indicating that greater representational capacity, not only useful features, contributes to the increase.

\section{Discussions}
\label{sec:discussion}

In this section we will discuss how our unified framework can be applied to learning settings beyond fully connected neural networks or simple task sequences. Several of these learning settings are closely related to \LLL in their setup and scope: multi-task learning, curriculum learning, few-shot learning, and convolutional networks. We will provide a brief discussion of their connections to our framework. This section concludes with a discussion on parallels with human learning. 



\subsection{Multi-Task and Curriculum Learning}
\label{sec:rd_curriculum}

Special cases of \LLL are multi-task learning \citep{caruana1997multitask} and curriculum learning \citep{bengio2009curriculum}. In multi-task learning, all $(T_{i}, D_{i})$ are provided together, allowing the model to be trained on all tasks at the same time. In curriculum learning, all data is similarly made available, but the problem is to identify the optimal order in which to train on data for the most efficient and effective learning. 
An example of an intuitive type of curriculum is to learn tasks from ``easy'' to ``hard'' \citep{elman1993learning}, similar to the way humans often learn new concepts.

\subsection{Few-shot Learning and Bongard Problems}
\label{sec:rd_bongard}

\begin{figure}[ht]
    \centering
    \includegraphics[width=0.8\textwidth]{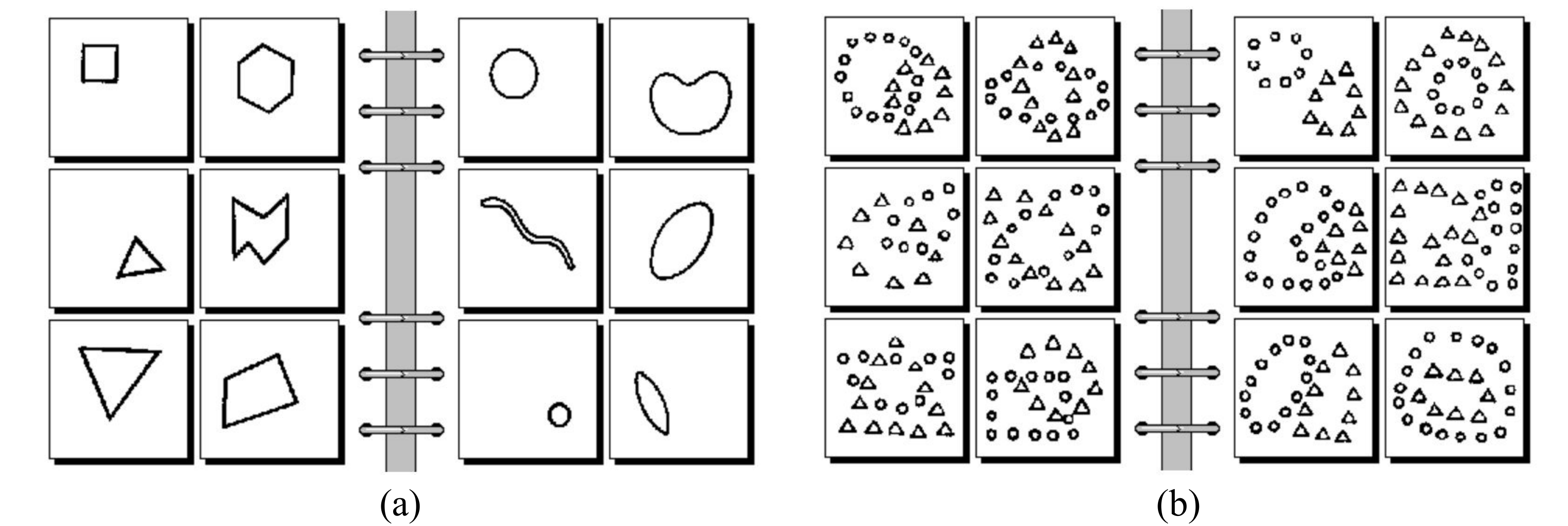}
    \caption{Two examples of BPs. (a) BP \#5 is simple, where the rule is polygonal shapes on the left and curvilinear shapes on the right. (b) BP \#99 is quite hard, where the rule is that larger shapes formed by connecting similar small shapes overlap vs. the larger shapes do not overlap. Such highly abstract rules are common among BPs. For difficult BPs, we find that for both humans and ML models, adding more training samples often does not help. Instead, it takes a ``stroke of insight'' or sufficiently similar previously acquired knowledge to solve it. 
    }
    \label{fig:BP_examples}
\end{figure}

Lifelong and curriculum learning have great potential to achieve few-shot learning of difficult concepts. By learning relevant easier concepts earlier with \LLL, later harder ones could be learned with many fewer examples via forward transfer of knowledge. 

Humans can often discover the underlying rules and patterns of complex concepts with only a few examples.  
We can use Bongard Problems (BPs) \citep{bongard1967problem}
to illustrate this\footnote{A collection of BPs can be found here: \url{http://www.foundalis.com/res/bps/bpidx.htm}.}.
A particular BP asks a human to recognize the underlying rule of classification with
only 6 examples on the left (say they are positive) and 6 examples on the right (negative samples).   
Two such BPs are shown in Figure~\ref{fig:BP_examples},
where (a) is a simple BP and (b) is a difficult BP for humans.
To solve BPs, humans often rely on highly abstract rules learned earlier in their lives (or in easier BPs). 
Some form of \LLL of many visual abstract concepts and shapes must likely have happened in humans so that they can solve BPs well with only 12 examples in total.  

\begin{figure}[ht]
    \centering
    \includegraphics[width=0.85\textwidth]{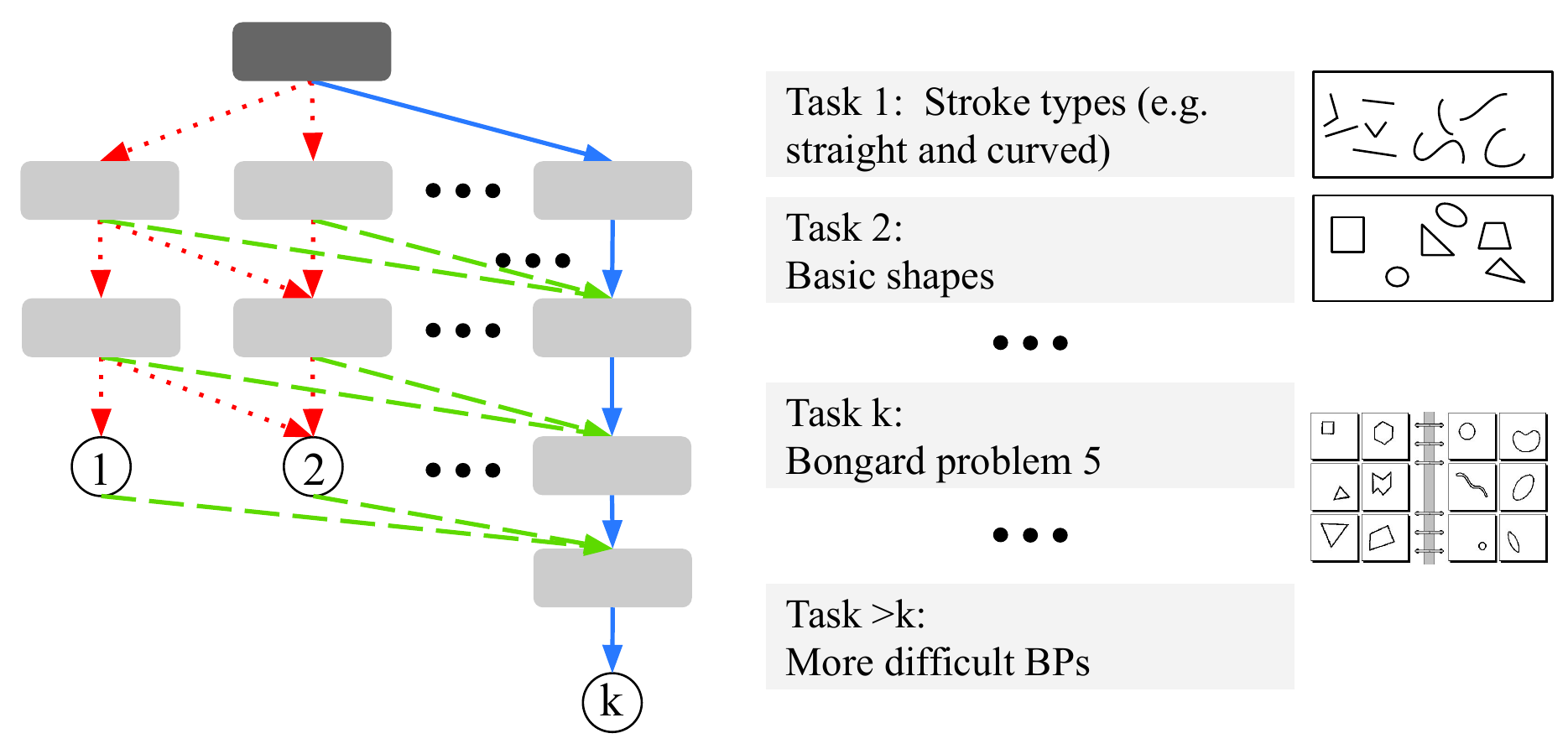}
    \caption{An illustrative example of curriculum learning to work up to solving Bongard problems with a small number of positive and negative examples. After learning to differentiate between basic strokes types, shapes, and possibly other early tasks, these representations including the output and hidden layers can be transferred to $T_{j}$ to possibly solving BPs such as BP\#5 (sharp angles vs. smooth curves) with a few examples. 
    This process can continue to allow us to solve a more complex visual reasoning problem with a few samples. Note that networks of subsequent tasks can have increasing depths in order to solve harder BPs after easier ones have been learned.}
    \label{fig:Curriculum_BP}
\end{figure}

We can outline a possible few-shot learning process to learn such BPs in our \LLL framework, in a curriculum-like fashion. 
First, we can train (or pre-train) on simpler tasks to recognize simpler shapes.
Next, harder and more complex visual shapes and BPs would be 
trained. 
This idea of using curriculum learning to ``working up to'' more difficult BPs is reflected in Figure~\ref{fig:Curriculum_BP}. 
In this illustrative example, we feed previous class labels (the output nodes) into nodes of new tasks to provide opportunity for forward transfer of knowledge. With this approach, BPs could be solved with fewer examples, as shown by \citet{yun2020deeper}. 
Note that in this illustrative example, the networks of subsequent tasks can have different and increasing depths in order to solve more difficult problems after easier problems are learned.  

To solve difficult BPs with a few examples (such as BP \#99 in Figure~\ref{fig:BP_examples}b), forward transfer of knowledge is crucial. If a person cannot solve BP \#99 (or other hard BPs) in several minutes, normally more training data would not help much.  Often ``a flash of inspiration'' or ``Aha!'' moment would suddenly occur and the problem is solved.  This is because a person has learned probably thousands or tens of thousands of concepts and abstract relations in his/her life, and the correct knowledge is suddenly combined to solve the hard BP.   

\subsection{Pre-Training Methods}
\label{sec:pretraining}

Pre-training is a ubiquitous process in deep learning \citep{devlinetal2019bert, girshick2014rich, wang2015unsupervised}. 
In computer vision (CV), state-of-the-art approaches often pre-train on a non-target task for which there is abundant data \citep{mahajan2018exploring}, such as the ImageNet dataset \citep{russakovsky2015imagenet}. Pre-trained CV models such as VGG-16 \citep{simonyan2014very} have been publicly released, allowing anyone to fine-tune their model to achieve otherwise infeasible performance on a target task. By reducing the need for target task data, pre-trained CV models have become a common approach to few-shot learning. \citep{ramalho2019empirical}.
In natural language processing, pre-trained language models have become popular \citep{conneau2017supervised, peters2018deep, devlinetal2019bert}. For example, BERT (Bidirectional Encoder Representations from Transformers) has shown that state-of-the-art performance can be attained on several different NLP tasks with one pre-trained model \citep{devlinetal2019bert}.

With pre-training, often a large neural network is first trained on one or more tasks with very large labeled training datasets.
The trained neural network would then be frozen, 
and a smaller network stacked on top of the large network and trained 
for improved performance on target tasks. From the perspective of our framework, the consolidation values of the smaller networks would be considered unfrozen for training.   

Pre-training can thus be viewed as certain weight consolidation policies in our \LLL framework. 
The main difference is that most of our consolidation algorithms in the last two sections apply to weights per tasks, or column-wise in our figures, while for pre-training, consolidation is layer-wise. An interesting research topic would thus be to consider designing consolidation policies which perform both layer- and column-wise consolidation based on the task at hand.

A natural way to fine-tune the consolidation policy when tuning on the target task is to consider how much to unfreeze each layer. Instead of simply tuning the output layer or tuning the entire network, we can interpolate between these two policies, to allow 
gradual unfreezing of layers in the network to best adapt to the target task 
without losing the generalization benefits of pre-training.  See \citep{erhan2010does} for an example of such directions.

\subsection{Convolutional Networks}
\label{sec:rd_cnns}


We can consider the application of our framework to convolutional neural networks (CNNs), not only fully connected feed-forward networks. Training on image tasks is often performed with resource-intensive batch learning, while our framework would allow for increased efficiency while being careful to maintain the high performance associated with batch learning. 

%

To adapt the consolidation mechanism to CNNs,
each $\pmb{b}_{i}$ can correspond to a filter rather than a single weight (as in a densely connected layer). A filter is essentially a set of weights which takes the representation from the previous layer (feature maps), and transforms it into a new feature map. Large $\pmb{b}$ on a filter means that it cannot be easily modified during learning. Next, we consider how to achieve the various \LLL properties in CNNs:

\begin{figure}[ht]
    \centering
    \includegraphics[width=0.55\textwidth]{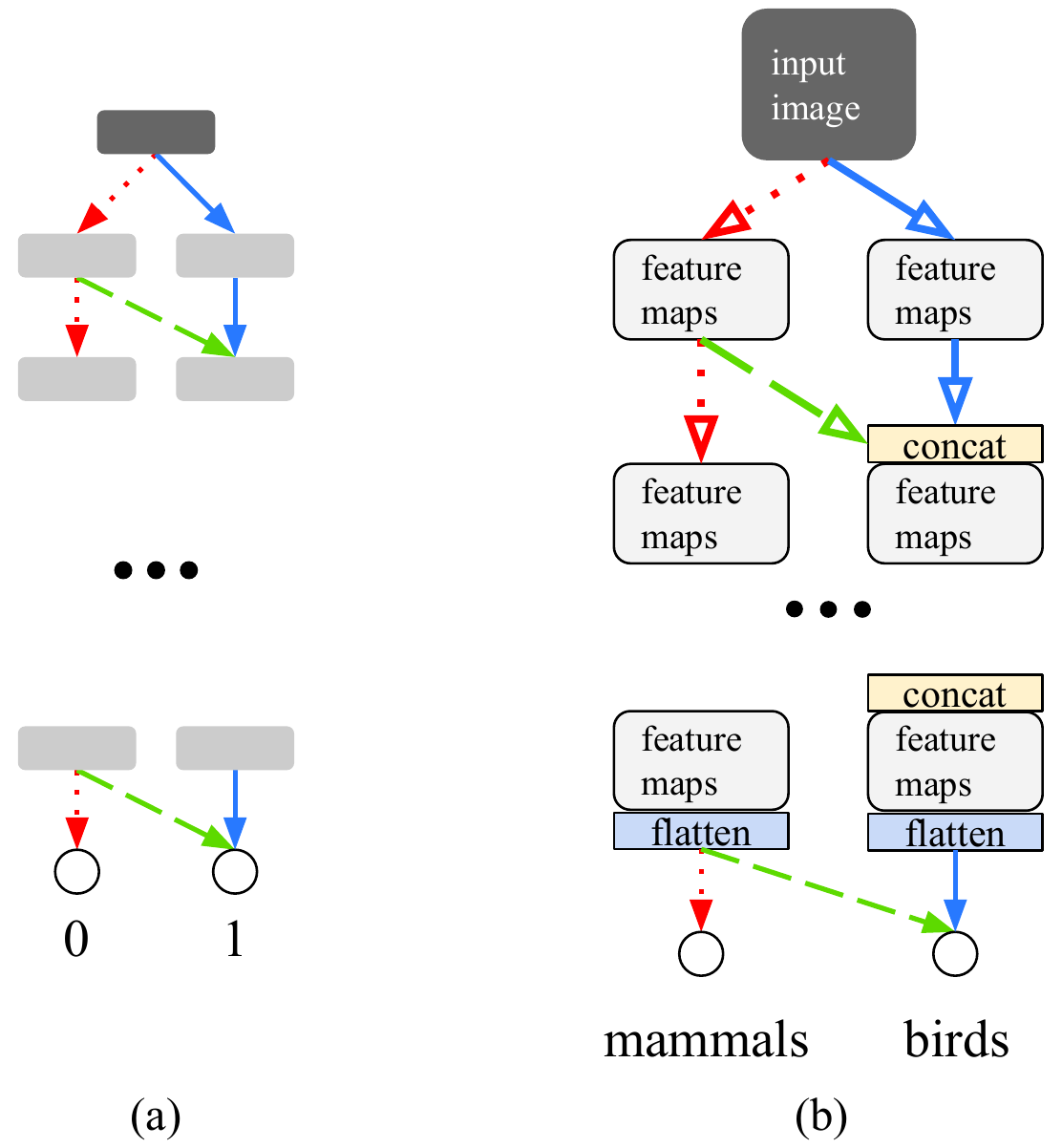}
    \caption{A simplified visualization of how our framework can be applied to convolutional networks. In (a) is a diagram of our consolidation mechanism and expansion being applied for learning two tasks with a fully connected network. In (b) is the corresponding CNN with the same high-level topology as (a) being applied to e.g. mammals for $T_{0}$ and birds for $T_{1}$. Links with unfilled arrow heads indicate filter weights, and links with filled arrow heads indicate weights between fully connected layers. Dotted (red) indicates large $\pmb{b}$ values, solid (blue) indicates $\pmb{b}$ values of 0, and dashed (green) indicates transfer links whose consolidation may depend on task similarity.}
    \label{fig:cnn}
\end{figure}

\paragraph{Continual learning of new tasks without forgetting} To extend the network for new tasks, we now add columns of convolutional filters, as reflected in Figure~\ref{fig:cnn}b (solid links).
For previous tasks not to be forgotten, their $\pmb{b} = \infty$
(see the red dotted links of Figure~\ref{fig:cnn}). 
For more difficult tasks, we can add a greater number of filters (analogous to a greater number of nodes). 


\paragraph{Forward transfer} The two techniques for encouraging forward transfer also extend to CNNs. First, instead of copying weights from previous tasks when similar to the new task, we can now copy over filter values. This can be done using similar ideas to those laid out in Section~\ref{sec:forwardtransfer}. Second, we can encourage forward transfer and avoid negative transfer through 
disconnecting the forward transfer links; 
essentially these filters are initialized to zero and the $\pmb{b} = \infty$, 
preventing negative transfer of knowledge from the previous task.

\subsection{Speculations on Parallels with Human Learning}
\label{sec:parallels}

\begin{wrapfigure}[34]{r}{0.4\textwidth}
	\centering
	\includegraphics[width=1\columnwidth]{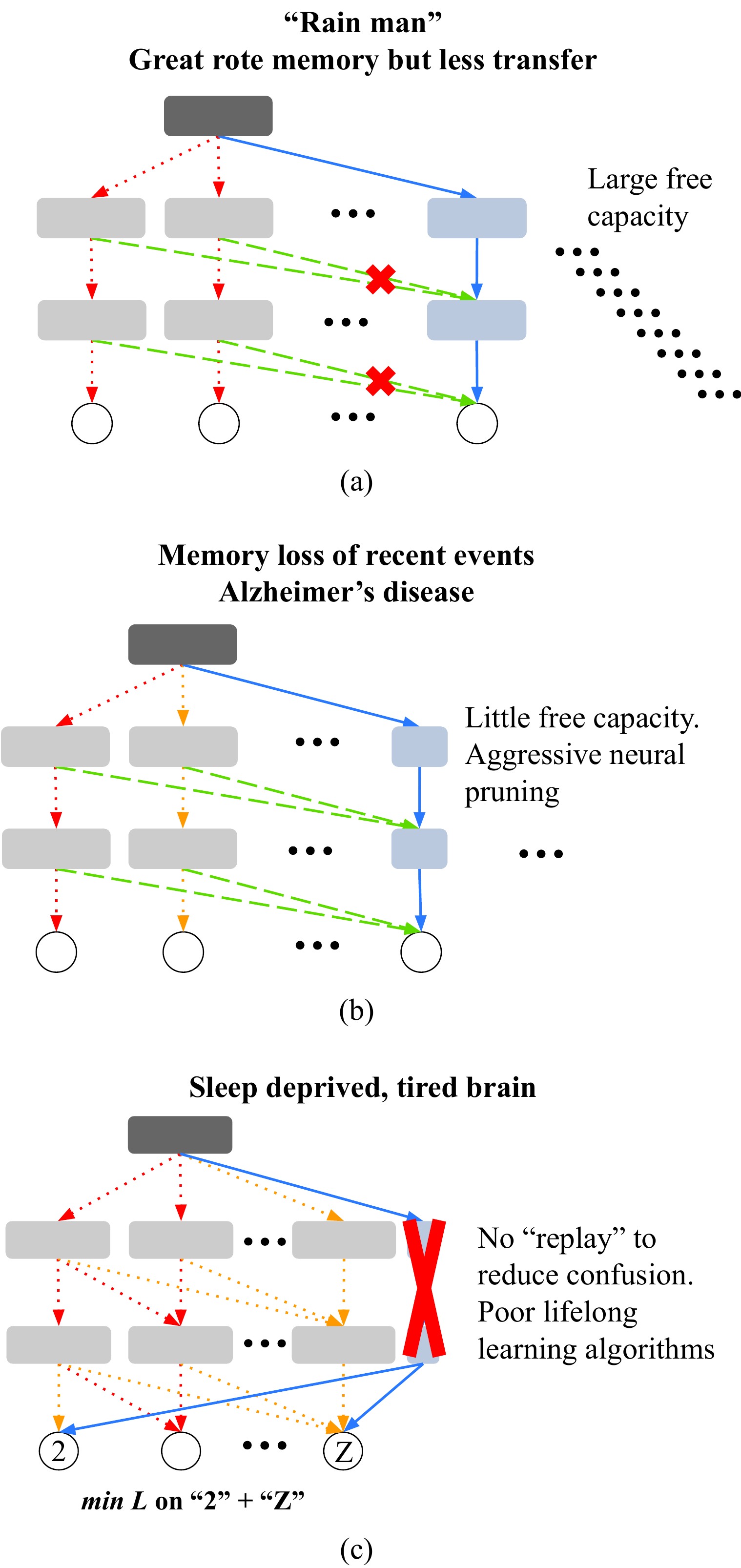}
	\caption{Illustrations of parallels between human learning behaviours and those that can be exhibited by our unified \LLL framework. Orange dotted links represent intermediate $\pmb{b}$ values.}
	\label{fig:parallels}
\end{wrapfigure}

As our framework focuses on controlling the flexibility of individual network weights, it is natural to ask how our approach lines up with the mammalian brain whose lifelong concept learning behaviours we are attempting to capture.
We will discuss briefly how certain behaviours can be exhibited by our framework with specific network sizes, training, and consolidation policies, 
and how it may translate to what happens in the human brain at a similar level of abstraction.

\paragraph{Resourceful and versatile} Ideally, human learning allows us to acquire a lot of knowledge, yet be flexible enough to adapt to new experiences and make meaningful connections between experiences.
This is analogous to our framework when everything works perfectly, including but not limited to: the network having an ample supply of free units for new tasks (Section~\ref{sec:continual}), using previous knowledge to learn new tasks (Section~\ref{sec:forwardtransfer}), not forgetting previous tasks while learning new ones (Section~\ref{sec:continual}), and using new knowledge to refine old skills (Section~\ref{sec:backwardtransfer}).

\paragraph{``Rain man''}
If $\pmb{b}$ values for previous tasks are very large to achieve non-forgetting (Section~\ref{sec:continual}), and no connections are made between previous and new tasks, knowledge transfer (Section~\ref{sec:forwardtransfer} and \ref{sec:backwardtransfer}) will likely not happen.  
This is reminiscent of Kim Peek, who was able to remember vast amounts of information, but performed poorly at abstraction-related tasks that require connecting unrelated skills and information and was the inspiration for the main character of the movie \textit{Rain Man} \citep{treffert2005inside}.
See Figure~\ref{fig:parallels}a for an illustration of our framework used to model similar behaviour.

\paragraph{Memory loss}
As shown in Section~\ref{sec:gracefulforgetting},
when availability of free units is limited, or worse, if units of the neural network are pruned away, graceful forgetting can be used to learn new tasks. 
In certain brain diseases, some ``tasks'' could be forgotten first. 
For example, 
an early stage of Alzheimer's disease 
(a highly complex neurological disease not fully understood)
is usually characterized by good memory on events years ago but poor on recent events \citep{tierney1996prediction}. This can be modeled by
large $\pmb{b}$ values for old tasks and small values for recent ones (shown as orange links in Figure~\ref{fig:parallels}b). 
Aggressive neuron pruning may also be happening (similar to node pruning in deep neural networks) to degrade the performance of tasks.

\paragraph{Sleep deprived} The human brain is suspected of performing important memory-related processes during sleep, and sleep deprivation is detrimental to memory performance \citep{walker2010sleep, killgore2010effects}. Confusion reduction and backward transfer are important stages of our proposed approach which use rehearsal (functionally similar to memory replay), where the model is exposed to samples from past tasks in order to perform fine-tuning to achieve various properties (Section~\ref{sec:confusionreduction} and \ref{sec:backwardtransfer}). Without these rehearsal-employing steps, the model may be less able to distinguish between samples of similar classes. Additionally, the ability to identify connections between newer tasks and older ones will be lost, so that potentially useful newly acquired skills cannot benefit older tasks. In addition, when the brain is ``tired'' and not well-rested, this may be analogous to poor optimization of policies in our framework (and in all ML algorithms). This is reflected in Figure~\ref{fig:parallels}c through the orange transfer and new task weights, which would reduce the ability for the new task to be learned well or efficiently make use of previous task knowledge. If optimization is not thorough in Equation~\ref{eqn:loss}, performance would be poor in all aspects of \LLL. 
See Figure~\ref{fig:parallels}c for an illustration.

\section{Conclusions}
\label{sec:conclusions}

In this work, we presented a conceptual unified framework for \LLL using one central mechanism based on consolidation.
We discussed how our approach can capture many important properties of lifelong learning of concepts, including non-forgetting, forward and backward transfer, confusion reduction, and so on, under one roof. Proof-of-concept results were reported to support the feasibility of these properties.
%
%
Rather than aiming for state-of-the-art results, this paper proposed research directions to help inform future \LLL research, including new algorithms and theoretical results. 
Last, we noted several similarities between our models with different training and consolidation policies and certain behaviors in human learning.

\acks{We are thankful for the constructive discussion and comments from lab members and colleagues on the many drafts of this work. We also acknowledge the support of the Natural Sciences and Engineering Research Council of Canada (NSERC) through the Discovery Grants Program. NSERC invests annually over \$1 billion in people, discovery and innovation.}


\newpage
\bibliography{main.bib}

\end{document}